\pdfoutput=1
\documentclass[11pt]{article}

\usepackage{naacl2021}
\usepackage{natbib}

\usepackage{times}
\usepackage{latexsym}
\usepackage{graphicx}
\usepackage{booktabs}
\usepackage[caption=false]{subfig}
\usepackage{colortbl}
\usepackage{enumitem}
\usepackage{color}
\usepackage{xcolor}
\usepackage{xspace}
\usepackage{multirow}
\usepackage[maxfloats=256]{morefloats}
\newcommand{\myparagraph}[1]{\noindent \textbf{#1}}

\newcommand{\Sref}[1]{\S\ref{#1}}

\newcommand{\fref}[1]{Figure~\ref{#1}}

\usepackage[T1]{fontenc}

\usepackage[utf8]{inputenc}

\usepackage{microtype}

\title{Modeling Framing in Immigration Discourse on Social Media}

\author{Julia Mendelsohn \\
  University of Michigan \\
  \texttt{juliame@umich.edu} \\\And
  Ceren Budak \\
  University of Michigan \\
  \texttt{cbudak@umich.edu} \\ \And
  David Jurgens \\
  University of Michigan \\
  \texttt{jurgens@umich.edu} \\}

\begin{document}
\maketitle
\begin{abstract}
The framing of political issues can influence policy and public opinion. Even though the public plays a key role in creating and spreading frames, little is known about how ordinary people on social media frame political issues. By creating a new dataset of immigration-related tweets labeled for multiple framing typologies from political communication theory, we develop supervised models to detect frames. We demonstrate how users' ideology and region impact framing choices, and how a message's framing influences audience responses. We find that the more commonly-used issue-generic frames obscure important ideological and regional patterns that are only revealed by immigration-specific frames. Furthermore, frames oriented towards human interests, culture, and politics are associated with higher user engagement. This large-scale analysis of a complex social and linguistic phenomenon contributes to both NLP and social science research.

\end{abstract}

\section{Introduction}


\textit{Framing} selects particular aspects of an issue and makes them salient in communicating a message \citep{Entman1993}. Framing can impact how people understand issues, attribute responsibility \citep{iyengar1991anyone}, and endorse possible solutions, thus having major implications for public opinion and policy decisions \citep{Chong2007}. While past work has studied framing by the news media and the political elite, little is known about how ordinary people frame political issues. Yet, framing by ordinary people can influence others' perspectives and may even shape elites' rhetoric \citep{RussellNeuman2014}. To shed light on this important topic, we focus on one issue---immigration---and develop a new methodology to computationally analyze its framing on Twitter. 

Our work highlights unique insights that social media data offers. The massive amount of available social media content enables us to compare framing strategies across countries and political ideologies. Furthermore, social media provides unique insights into how messages resonate with audiences through interactive signals such as retweets and favorites. By jointly analyzing the production and reception of frames on Twitter, we provide an in-depth analysis of immigration framing by and on the public. 

Political communications research has identified numerous typologies of frames, such as \textit{issue-generic policy}, \textit{immigration-specific}, and \textit{narrative}. Each of these frame types can significantly shape the audience's perceptions of an issue \citep{iyengar1991anyone,Chong2007,Lecheler2015}, but prior NLP work seeking to detect frames in mass media (e.g. \citealp{Card2016,Field2018,kwak2020systematic}) has largely been limited to a single \textit{issue-generic policy} typology. Multiple dimensions of framing must be considered in order to better understand the structure of immigration discourse and its effect on public opinion and attitudes. We thus create a novel dataset of immigration-related tweets containing labels for each typology to facilitate more nuanced computational analyses of framing.

This work combines political communication theory with NLP to model multiple framing strategies and analyze how the public on Twitter frames immigration. Our contributions are as follows: 
(1) We create a novel dataset of immigration-related tweets labeled for  {issue-generic policy}, {immigration-specific}, and {narrative} frames. (2) We develop and evaluate multiple methods to detect each type of frame. (3) We illustrate how a message's framing is influenced by its author's ideology and country. (4) We show how a message's framing affects its audience by analyzing favoriting and retweeting behaviors.
%
%
Finally, our work highlights the need to consider multiple framing typologies and their effects.  

\section{Framing in the Media}

Framing serves four functions: (i) defining problems, (ii) diagnosing causes, (ii) making evaluative judgments, and  (iv) suggesting solutions \citep{Entman1993}. Framing impacts what people notice about an issue, 
making it a key mechanism by which a text influences its audience. 

\begin{table*}[t]
\centering

\resizebox{\textwidth}{!}{%
\begin{tabular}{@{}lll@{}}

\multicolumn{1}{c}{\textbf{Frame Type}} & \multicolumn{1}{c}{\textbf{Frame}} & \multicolumn{1}{c}{\textbf{Description}} \\ \hline
Issue-Generic & Economic & Financial implications of an issue \\ \rowcolor{gray!15}
\cellcolor{white} Policy & Capacity \& Resources & The availability or lack of time, physical, human, or financial resources \\
 & Morality \& Ethics & Perspectives compelled by religion or secular sense of ethics or social responsibility \\ \rowcolor{gray!15}
 & Fairness \& Equality & The (in)equality with which laws, punishments, rewards, resources are distributed \\
 & \begin{tabular}[c]{@{}l@{}}Legality, Constitutionality\\ \& Jurisdiction\end{tabular} & \begin{tabular}[c]{@{}l@{}}Court cases and existing laws that regulate policies; constitutional interpretation;\\ legal processes such as seeking asylum or obtaining citizenship; jurisdiction\end{tabular} \\ \rowcolor{gray!15}
 & Crime \& Punishment & The violation of policies in practice and the consequences of those violations \\
 & Security \& Defense & Any threat to a person, group, or nation and defenses taken to avoid that threat \\ \rowcolor{gray!15}
 & Health \& Safety & Health and safety outcomes of a policy issue, discussions of health care \\
 & Quality of Life & Effects on people's wealth, mobility, daily routines, community life, happiness, etc. \\ \rowcolor{gray!15}
 & Cultural Identity & Social norms, trends, values, and customs; integration/assimilation efforts \\
 & Public Sentiment & General social attitudes, protests, polling, interest groups, public passage of laws \\ \rowcolor{gray!15}
 & \begin{tabular}[c]{@{}l@{}}Political Factors \& \\ Implications\end{tabular} & \begin{tabular}[c]{@{}l@{}}Focus on politicians, political parties, governing bodies, political campaigns \\ and debates; discussions of elections and voting\end{tabular} \\
 & \begin{tabular}[c]{@{}l@{}}Policy Prescription \&\\ Evaluation\end{tabular} & Discussions of existing or proposed policies and their effectiveness \\ \rowcolor{gray!15}
 & \begin{tabular}[c]{@{}l@{}}External Regulation \& \\ Reputation\end{tabular} & \begin{tabular}[c]{@{}l@{}}Relations between nations or states/provinces; agreements between governments; \\ perceptions of one nation/state by another\end{tabular} \\ \hline
Immigration & Victim: Global Economy & Immigrants are victims of global poverty, underdevelopment and inequality \\ \rowcolor{gray!15}
\cellcolor{white} Specific & Victim: Humanitarian & Immigrants experience economic, social, and political suffering and hardships \\
 & Victim: War & Focus on war and violent conflict as reason for immigration \\ \rowcolor{gray!15}
 & Victim: Discrimination & Immigrants are victims of racism, xenophobia, and religion-based discrimination \\
 & Hero: Cultural Diversity & Highlights positive aspects of differences that immigrants bring to society \\ \rowcolor{gray!15}
 & Hero: Integration & Immigrants successfully adapt and fit into their host society \\
 & Hero: Worker & Immigrants contribute to economic prosperity and are an important source of labor \\ \rowcolor{gray!15}
 & Threat: Jobs & Immigrants take nonimmigrants' jobs or lower their wages \\
 & Threat: Public Order & Immigrants threaten public safety by being breaking the law or spreading disease \\ \rowcolor{gray!15}
 & Threat: Fiscal & Immigrants abuse social service programs and are a burden on resources \\
 & Threat: National Cohesion & Immigrants' cultural differences are a threat to national unity and social harmony \\ \hline \rowcolor{gray!15}
\cellcolor{white} Narrative & Episodic & Message provides concrete information about on specific people, places, or events \\
 & Thematic & Message is more abstract, placing stories in broader political and social contexts \\ 
\end{tabular}%
}
\caption{List of all issue-generic policy \citep{Boydstun2013}, immigration-specific \citep{benson2013shaping,Hovden2019}, and narrative \citep{iyengar1991anyone} frames with brief descriptions.}
\label{tab:frame_descriptions}
\end{table*}

\myparagraph{Framing Typologies} 
%
We draw upon distinct typologies of frames that can be  applied to the issue  of immigration: (1) issue-specific, which identify aspects of a particular issue, or (2) issue-generic, which appear across a variety of issues and  facilitate cross-issue comparison \citep{DeVreese2005}. 

Issue-generic frames include policy frames that focus on aspects of issues important for policy-making, such as economic consequences or fairness and equality \citep{Boydstun2013}. Other generic frames focus on a text's narrative; news articles use both \textit{episodic} frames, which highlight specific events or individuals, and \textit{thematic} frames, which place issues within a broader social context. The use of \textit{episodic} versus \textit{thematic} frames can influence the audience's attitudes. For example, \textit{episodic} frames lead audiences to attribute responsibility for issues such as poverty to individual citizens while \textit{thematic} frames lead them to hold the government responsible \citep{iyengar1991anyone}.  

Issue-specific frames for immigration focus on the portrayal of immigrants. Our analysis uses \citet{benson2013shaping}'s set of issue-specific frames, which represent immigrants as heroes (cultural diversity, integration, good workers), victims (humanitarian, global economy, discrimination), and threats (to jobs, public order, taxpayers, cultural values).

Both issue-specific and generic frames provide unique insights but present advantages and drawbacks. While issue-specific frames analysis are specific and detailed, they are hard to generalize and replicate across studies, which is a key advantage for generic frames \citep{DeVreese2005}. 

\myparagraph{Framing effects}
Studies of framing typically focus on either \textit{frame-building} or \textit{frame-setting}~\citep{Scheufele1999, DeVreese2005}. \textit{Frame-building} is the process by which external factors, such as a journalist's ideology or economic pressures, influence what frames are used; frame-building studies thus treat framing as the dependent variable. Frame-setting studies treat frames as independent variables that impact how an audience interprets and evaluates issues.

Prior analyses of frame-building in immigration news highlight region and ideology as particularly important factors. Right-leaning media from conservative regions are more likely to frame immigrants as intruders~\citep{VanGorp2005}, and as threats to the economy and public safety~\citep{Fryberg2012}. Framing also differs across countries; while the US press emphasizes public order, discrimination, and humanitarian concerns, the French press more frequently frames immigrants as victims of global inequality \citep{benson2013shaping}.

Frame-setting has also been studied in the context of immigration. For example, experimental work has shown that frames eliciting angry or enthusiastic emotions impact participants' opinions on immigration \citep{Lecheler2015}. While past work has analyzed linguistic framing in Twitter immigration discourse \citep[e.g.,][]{DeSaintLaurent2020}, little is known about how such framing affects users' interactive behaviors such as resharing content, which is a key objective of frame setting.

\section{Computational Approaches to Framing}

Because many people now generate and consume political content on social media, scholars have increasingly used automated techniques to study framing on social media. 

Large-scale research of framing on Twitter has commonly focused on unsupervised approaches. (e.g., \citealp{RussellNeuman2014, Meraz2013,DeSaintLaurent2020}). Such approaches, including those focused on hashtag analysis, can reveal interesting framing patterns. 
For instance, \citet{Siapera2018} shows that frame usage varies across events.
Similarly, topic models have been used to compare ``refugee crisis" media discourses across the European countries \citep{Heidenreich2019}, and to uncover differences in attitudes towards migrants \citep{Hartnett2019}.
Although lexicon analysis and topic models can provide insights about immigration discourse, here, we adopt a supervised approach to ground our work in framing research and to enable robust evaluation. 


We draw inspiration from a growing body of NLP research that uses supervised approaches to detect issue-generic policy frames in news articles, a task popularized by the Media Frames Corpus~\citep{Card2015}, which contains issue-generic policy frame labels for articles across several issues~\citep{Boydstun2013}. Using this corpus, prior work has detected frames with techniques including logistic regression \citep{Card2016}, recurrent neural networks \citep{Naderi2017}, lexicon induction \citep{Field2018}, and fine-tuning pretrained language models \citep{Khanehzar2019,kwak2020systematic}. \citet{roy2020weakly} further extracted subcategories of issue-generic policy frames in newspaper coverage using a weakly-supervised approach. Finally, issue-generic frames have also been computationally studied in other media, including online fora and politicians' tweets \citep{Johnson2017,Hartmann2019a}. We build upon this literature by incorporating additional frame typologies that reflect
important dimensions of media discourse with real-world consequences \citep{iyengar1991anyone,gross2008,Eberl2018}. Beyond detecting frames, we computationally analyze frame-building and frame-setting among social media users; though well-studied in traditional news media, little is known about how social media users frame immigration or its effects \citep{Eberl2018}.

Noting that issue-generic policy frames obscure important linguistic differences, several works studied issue-specific frames in news media for issues such as missile defense and gun violence \citep{Morstatter2018,liu2019detecting}. We extend issue-specific frame analyses to immigration by adopting an immigration-specific typology developed by political communication scholars \cite{benson2013shaping}. 


In contrast to prior NLP work focused on traditional media or political elites \citep{Johnson2017,Field2018}, we highlight the role that social media publics play in generating and propagating frames. Furthermore, we provide a new computational model of narrative framing \cite{iyengar1991anyone}, that together with models for issue-generic policy and issue-specific frames, provides complementary views on the framing of immigration. Finally, our large-scale analysis of frame-setting illustrates the potential for using NLP to understand how a message's framing shapes its audience behavior.

\section{Data}

We first collect a large dataset of immigration-related tweets, and then annotate a subset of this full dataset for multiple types of frames. 

\myparagraph{Data Collection} 
We extract all English-language tweets in 2018 and 2019 from the Twitter Decahose containing at least one of the following terms: \textit{immigration}, \textit{immigrant(s)}, \textit{emigration}, \textit{emigrant(s)}, \textit{migration}, \textit{migrant(s)}, \textit{illegal alien(s)}, \textit{illegals}, and \textit{undocumented}\footnote{We obtained this list by starting with the seed terms \textit{immigrants}, \textit{immigration}, and \textit{illegal aliens}. We then added the remaining terms by manually inspecting and filtering nearby words in pretrained GloVe and Word2Vec vector spaces.}. We focus on content creation and thus exclude retweets from our dataset, though we consider retweeting rates when analyzing the social influence of different frames. We further restrict our dataset to tweets whose authors are identified as being located in the United States (US), United Kingdom (GB), and European Union (EU) by an existing location inference tool \citep{compton2014geotagging}.
To compare framing across political ideologies, we obtain ideal point estimates for nearly two-thirds of US-based users with \citet{Barbera2015a}'s Bayesian Spatial Following model. Our full dataset contains over 2.66 million tweets, 86.2\% of which are from the United States, 10.4\% from the United Kingdom, and 3.4\% from the European Union. 


\myparagraph{Data Annotation}
Tweets are annotated using three frame typologies: (i) issue-generic policy, (ii) immigration-specific, and (iii) {narrative} frames, where a tweet may use multiple frames simultaneously. We use \citet{Boydstun2013}'s Policy Frames Codebook to formulate our initial guidelines to code for policy frames. We use \citet{benson2013shaping}'s immigration-specific frames, but follow \citet{Hovden2019} in including an additional category for framing immigrants as victims of war. Finally, we code for narrative frames using definitions from \citet{iyengar1991anyone}. All frames and descriptions can be found in Table \ref{tab:frame_descriptions}, with a complete codebook in Supplementary Materials. Because annotation guidelines from prior work focus on elite communications, we first adjusted our codebook to address challenges posed by Twitter content. Changes were made based on feedback from four trained annotators who labeled 360 tweets from 2018, split between the EU, GB, and US. 

Even for humans, identifying frames in tweets is a difficult task. Defining the boundaries of what constitutes a message is not trivial. Beyond the text, frames could be identified in hashtags, images, videos, and content from linked pages. Furthermore, tweets are often replies to other users or part of a larger thread. This additional context may influence an issue's framing. For simplicity, we treat each tweet as a standalone message and label frames based only on the text (including hashtags).

Unlike news stories, where frames are clearly cued, tweets often implicitly allude to frames due to character limitations. For example, a tweet expressing desire to ``drive immigrants out" with no additional context may suggest a criminal frame, but criminality is not explicit. To minimize errors, we avoid making assumptions about intended meaning and interpret all messages literally.

Training, development, and test data were annotated using two procedures after four annotators completed four rounds of training. The dataset contains equal numbers of tweets from the EU, UK, and US. Training data was singly annotated and includes 3,600 tweets, while the development and test sets each contain 450 tweets (10\% of the full dataset) and were consensus-coded by pairs of trained annotators.
We opt for this two-tier approach due to (i) the inherent difficulty of the task\footnote{For example, in identifying just the primary issue-generic frame of a document, the Media Frames corpus attained an Krippendorff's $\alpha$=$\sim$0.6 \citep[][Fig.~4]{Card2015}, whereas we ask annotators to identify \textit{all} frames across three typologies. } and (ii) the need to maximize diversity seen in training. During annotator training, pilot studies attained moderate agreement, suggesting that to attain high-reliability, consensus coding with adjudication would be needed \cite{krippendorff2013content}, which comes at a cost of substantially increased time. Because a large dataset of unique, singly-coded documents is preferable to a small dataset of documents coded by multiple annotators for text classification~\citep{barbera2019automated}, we decided to increase corpus diversity in the training data by singly-annotating, at the expense of potentially noisier annotation, and to consensus code all evaluation data. On the double annotated data, annotators attained Krippendorff's $\alpha$=0.45. Additional details are provided in Supplementary Material (\Sref{app:agreement}, Figures \ref{fig:agreement_julia} and \ref{fig:agreement_consensus}). 



\myparagraph{Results}
We observe differences across frame typologies in coverage rates within the annotated data set. While 84\% of tweets are labeled with at least one {issue-generic policy} frame and 85\% with at least one {narrative} frame, only 51\% are labeled with at least one {issue-specific} frame. This difference is due to immigration-specific frames being more narrowly-defined, as they require explicit judgment of immigrants as heroes, victims, or threats. Further details about frame distributions in our annotations can be found in Supplementary Material (\Sref{app:distributions}, Figure \ref{fig:distribution}).

While the precision of {issue-specific} frames can reveal patterns otherwise obscured by the broader issue-generic frames, this lack of coverage presents two challenges: 1) automated detection is more challenging given this sparsity and 2) analyses of {issue-specific} frames do not capture a large portion of immigration-related discourse. 
By incorporating multiple framing strategies, we leverage both the coverage of issue-generic frames and the precision and interpretability of issue-specific frames.


\section{Frame Detection}

We formulate frame detection as a multilabel classification problem for each of the three typologies, using our dataset to train supervised models. 

\myparagraph{Experimental Setup}
%
Our proposed model is a RoBERTa model \citep{liu2019roberta} trained using binary cross-entropy on the CLS token. We consider both (i) a model trained using the \texttt{roberta-base} parameters and (ii) a second model that has first been fine-tuned on our full set of immigration tweets using masked-language modeling. Fine tuning was performed for 60 epochs. In both models, early stopping is used to avoid overfitting.  Models are compared with two baselines:  random prediction, and logistic regression with unigram and bigram features. Each model was trained five times with different random seeds and we report bootstrapped mean performance.


\begin{table}[]
\centering
\resizebox{\linewidth}{!}{
\begin{tabular}{ cccc }

\textbf{Random} & \textbf{LogReg} & \textbf{RoBERTa} & \textbf{FT RoBERTa} \\
\hline
 0.193 & 0.296 & 0.611 & 0.657 \\ 
\end{tabular}%
}
\caption{F1 scores  on  the  test  set  for  all models,  calculated  as  an  (unweighted)  average  over all  frames  and  initialization  seeds. The  fine-tuned (FT) RoBERTa model improvements over all models are significant at p$<$0.001 (McNemar's test). 
}
\label{tab:model_f1}
\end{table}

\begin{table}[t]
\centering
\resizebox{\linewidth}{!}{
\begin{tabular}{ r cccc}
\textbf{Frame Type} & \textbf{Precision} & \textbf{Recall} & \textbf{F1 Score} & \textbf{LRAP}\\ \hline
Issue-Generic Policy & 0.727 & 0.721 & 0.711 & 0.750\\
Issue-Specific & 0.593 & 0.531 & 0.552 & 0.806\\
Narrative & 0.757 & 0.887 & 0.808 & 0.894\\
\end{tabular}
}
\caption{Test set performance on each frame typology.}
\label{tab:eval_frame_type}
\end{table}

\myparagraph{Results}
The fine-tuned RoBERTa model significantly outperforms all baselines (Table \ref{tab:model_f1}). RoBERTa has the most substantial gains over logistic regression for low-frequency frames (Supplementary Material \Sref{app:model-performance}, \fref{fig:support_f1}). These gains for rare frames are essential for analyzing immigration discourse on social media in order to capture diverse perspectives and arguments. 

Table~\ref{tab:eval_frame_type} shows several evaluation metrics separated by frame type. Precision, recall, and F1 are calculated as unweighted averages over all frames belonging to each category. Overall, issue-generic policy and narrative frames can be detected more effectively than issue-specific frames. This difference reflects that issue-specific frames were sparser in the training data, but also that detecting these frames is inherently more challenging because it requires jointly reasoning about immigration-related topics and how these topics affect immigrants. For example, tweets about immigrants committing crimes and tweets about hate crimes committed against immigrants have distinct issue-specific frames (\textit{threat: public order} and \textit{victim: discrimination}), even though these texts can be linguistically quite similar. 
Given some thematic similarities between typologies, we tested an additional model that jointly predicted frames from all three typologies using the fine-tuned RoBERTa model; however, the resulting model offered worse performance than any single-typology model, suggesting minimal benefits of cross-typology learning. 
Supplementary Section \ref{app:model-performance} contains additional model performance analyses by frame and region. 
\begin{figure}[h!]
    \centering
    \includegraphics[width=\linewidth]{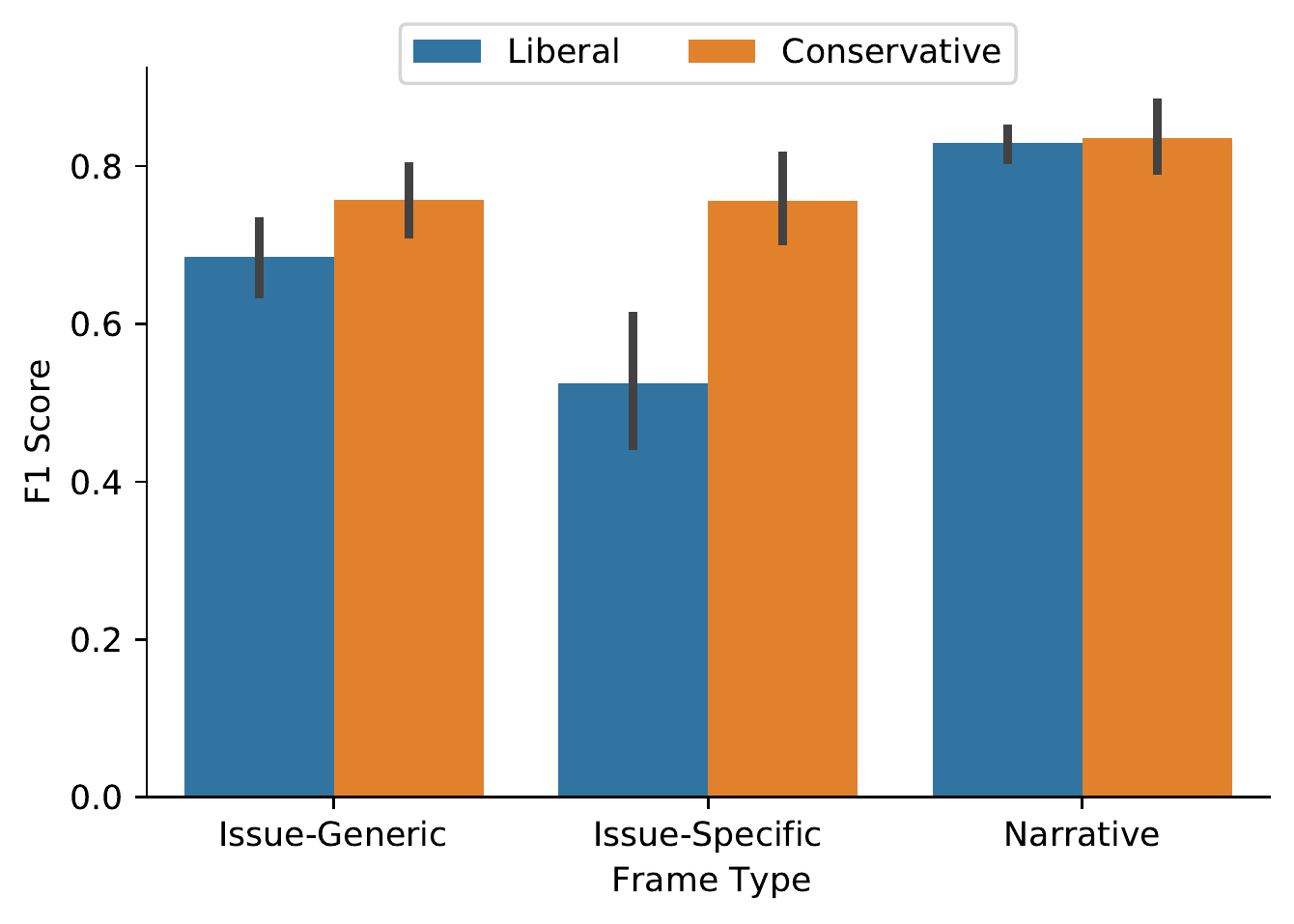}
    \caption{Average F1 scores on combined dev/test set separated by US authors' ideologies. Issue-generic policy and issue-specific frame detection models have higher performance for conservatives than liberals.}
    \label{fig:ideology_f1}
\end{figure}

\begin{table*}[]
\centering
\resizebox{\textwidth}{!}{%
\begin{tabular}{@{}lll@{}}

\textbf{Error Type} & \textbf{Description} & \textbf{Example}  \\ \hline
Plausible interpretation & \begin{tabular}[c]{@{}l@{}}These instances highlight the challenges of annotation;\\ there are convincing arguments that model's predicted \\ frames can be appropriate labels.\end{tabular} & \begin{tabular}[c]{@{}l@{}}Interestingly, the criteria to which immigrants would be held would \\ not be met by a large number of the ‘British’ people either. \\
\textit{Model erroneously predicted Policy}
\end{tabular} \\ \rowcolor{gray!15}
\begin{tabular}[c]{@{}l@{}}Inferring frames not \\ explicitly cued in text\end{tabular} & \begin{tabular}[c]{@{}l@{}}Model predicts frames that may capture an author's intention \\ but without sufficient evidence from the text\end{tabular} & 
\begin{tabular}[c]{@{}l@{}}Stop immigration \\ \textit{Model erroneously predicted Threat: Public Order}\end{tabular} \\
\begin{tabular}[c]{@{}l@{}}Missing necessary \\ contextual knowledge\end{tabular} & \begin{tabular}[c]{@{}l@{}}Some frames are directly cued by lexical items \\ (e.g. politicians' names cue Political frame), but model \\ lacks real-world knowledge required to identify these frames\end{tabular} & \begin{tabular}[c]{@{}l@{}}@EricTrump Eric I have been alive longer than your immigrant\\  mother in law and you. I paid more in taxes than you did and \\ your immigrant mother in law combined... \\ 
\textit{Model missed Political frame}
\end{tabular}  \\ \rowcolor{gray!15}
\begin{tabular}[c]{@{}l@{}}Overgeneralizing \\ highly-correlated features\end{tabular} & \begin{tabular}[c]{@{}l@{}}Many words and phrases do not directly cue frames, but are \\ highly-correlated. The model makes erroneous predictions \\ when such features are used in different contexts (e.g. violence \\ against immigrants, rather than immigrants being violent)\end{tabular} & \begin{tabular}[c]{@{}l@{}}Lunaria’s figures from 2018 recorded 12 shootings, two murders\\ and 33 physical assaults against migrants in the first two months \\ since Salvini entered government.
\\ \textit{Model missed Victim: Humanitarian frame}

\end{tabular} \\
Pronoun ambiguity & \begin{tabular}[c]{@{}l@{}}Coreference resolution is often not possible and annotators avoided \\ making assumptions to resolve ambiguities. For example, "you" \\ can be used to discuss individuals' experiences (episodic) but its \\ impersonal sense can be in broad generalizations (thematic).\end{tabular} & \begin{tabular}[c]{@{}l@{}}It's worse when you have immigrant parents who don't speak\\ the language cause you have to deal with all the paperwork, \\ be the translator for them whenever they go (...)\\ its tiring but someone has to \\ 
\textit{Model predicted Episodic but referent is unclear}
\end{tabular} \\ 
\end{tabular}%
}
\caption{Types of common errors in frame prediction along with brief descriptions and examples.}
\label{tab:errors}
\end{table*}

\myparagraph{Hegemonic Framing} 
Conservative media's framing of political issues is known to be more consistent, coordinated, and hegemonic than mainstream media, which has been vital to the success of the American conservative movement \citep{hemmer2016messengers,speakman2020}. If the same pattern holds for social media, we would expect automated frame detection to achieve higher performance on conservative tweets due to more linguistic regularities across messages. Indeed, we find that issue-generic and issue-specific classifiers achieve higher F1 scores on tweets written by conservative authors compared to liberal authors (Figure \ref{fig:ideology_f1}), even though there are fewer conservative tweets in the training data (334 conservative vs 385 liberal tweets). Higher model performance on conservative tweets suggests that, like political and media elites, conservatives on social media are more consistent than liberals in their linguistic framing of immigration.

\myparagraph{Error Analysis} We identify classification errors by qualitatively analyzing a random sample of 200 tweets that misclassified at least one frame. Table \ref{tab:errors} shows the most common categories of errors.

\section{Frame-Building Analysis}

In writing about an issue, individuals are known to select particular frames---a process known as frame-building---based on numerous factors, such as exposure to politicians' rhetoric  or their own identity \citep{Scheufele1999}. Here, we focus on two specific identity attributes affecting  frame building: (i) political ideology and (ii) country/region.

The political, social, and historical contexts of an one's nation-state can impact how they frame immigration \citep{Helbling2014}. Immigration has a long history in the USA relative to Europe, and former European colonial powers (e.g. the UK) have longer immigration histories than other countries (e.g. Norway) \citep{thorb2015,Eberl2018}. Cross-country variation in news framing also arise from differences in immigration policies \citep{Helbling2014,Lawlor2015}, media systems \citep{thorb2015}, journalistic norms \citep{Papacharissi2008}, geographic proximity to immigrant populations or points of entry \citep{Grimm2011,Fryberg2012}, and immigrants' race/ethnicity \citep{Grimm2011}. At the same time, increased globalization may result in a uniform transnational immigration discourse \citep{Helbling2014}. Framing variations across countries has implications for government policies and initiatives, particularly in determining what solutions could be applied internationally or tailored to each country \citep{Caviedes2015}.

Prior studies on the role of ideology in frame-building have focused on the newspapers or political movements, showing patterns in frames like morality and security by political affiliation in European immigration discourse \cite{Helbling2014,hogan2015} or in use of economic frames by American newspapers \cite{Fryberg2012,Abrajano2017}.
%
However, it remains unclear whether these patterns observed for elite groups can generalize to the effect of individual people's political dispositions.

\begin{figure}[t!]
    \centering
    \includegraphics[width=\linewidth]{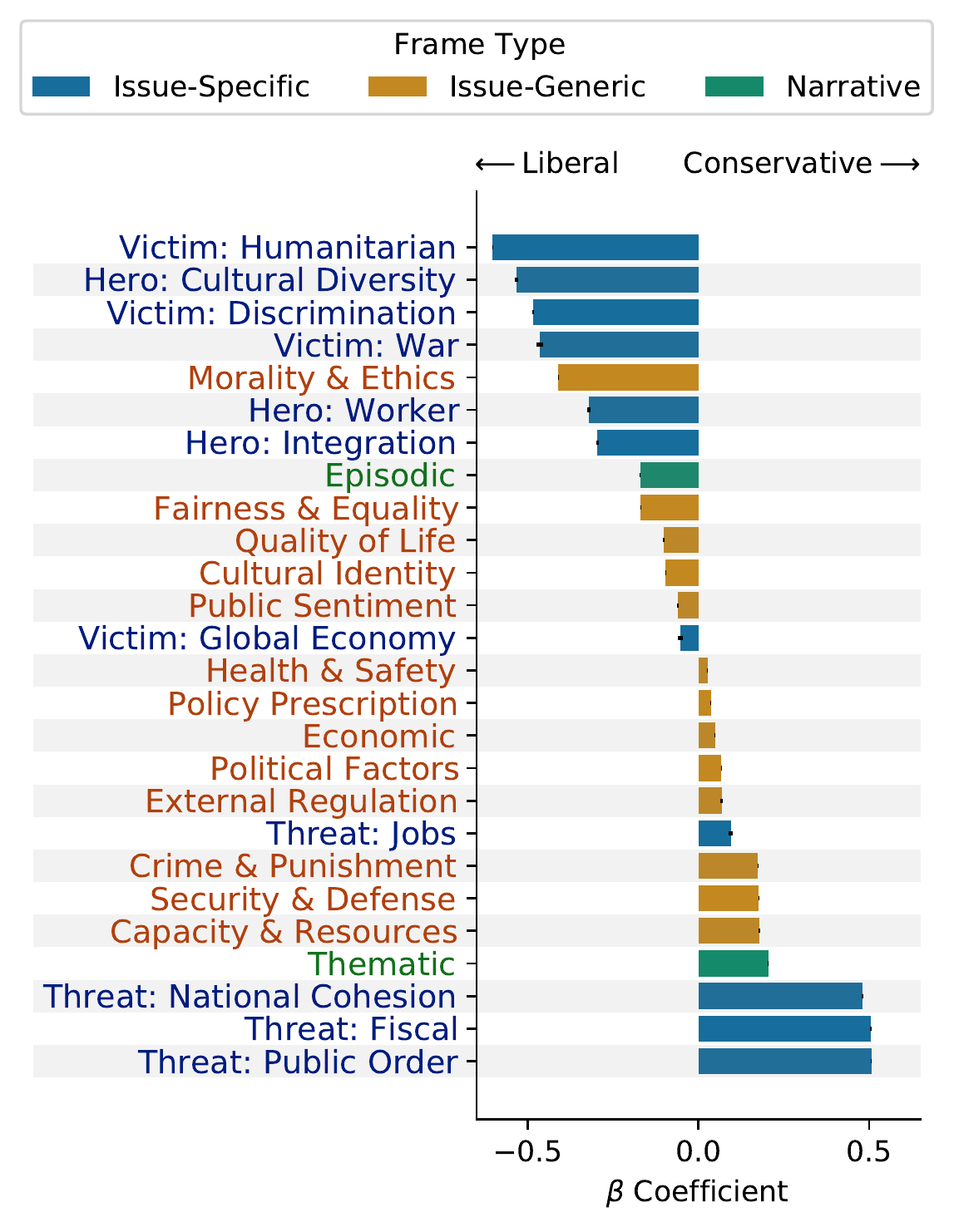}
    \caption{Logistic regression coefficients of political ideology in predicting each frame. Positive (negative) values correspond to more conservative (liberal) ideology.
    Only frames associated with ideology after Holm-Bonferroni correction with $p < 0.01$ are included.}
    \label{fig:ideology}
\end{figure}


\myparagraph{Experimental Setup}
We detect frames for all 2.6M immigration-related tweets using the fine-tuned RoBERTa model with the best-performing seed on development data. Using this labeled data, we estimate the effects of region and ideology by fitting separate mixed-effects logistic regression models to predict the presence or absence of each frame. We treat region (US, UK, and EU) as a categorical variable, with US as the reference level. 
Ideology is estimated using the method of \citet{Barbera2015a}, which is based on users' connections to US political elites; as such, we restrict our analysis of ideology to only tweets from the United States.

To account for exogenous events that may impact framing, we include nested random effects for year, month, and date. We further control for user characteristics (e.g. the author's follower count, friends count, verified status and number of prior tweets) as well as other tweet characteristics (e.g. tweet length, if a tweet is a reply, and whether the tweet contains hashtags, URLs, or mentions of other users). We apply Holm-Bonferroni corrections on p-values before significance testing to account for multiple hypothesis testing.

\myparagraph{Ideology}
Ideology is strongly predictive of framing strategies in all three categories, as shown in \fref{fig:ideology}. 
Our results reveal three broad themes. 

First, prior work has argued that liberals and conservatives adhere to different moral foundations, with conservatives being more sensitive to in-group/loyalty and authority than liberals, who are more sensitive to care and fairness \citep{graham2009liberals}. Our results agree with this argument. Liberals are more likely to frame immigration as a fairness and morality issue, and immigrants as victims of discrimination and inhumane policies.  
More conservative authors, on the other hand, focus on frames with implications for the in-group. They express concerns about 1) immigrants imposing a burden on taxpayers and governmental programs and 2) immigrants being criminals and threats to public safety. We qualitatively observe three distinct, though unsubstantiated, conservative claims contributing to the latter: (i.) Immigrants commit violent crimes \citep{light2018does}, (ii.) Undocumented immigrants illegally vote in US elections \citep{Smith148,udani2018immigrant}, and (iii.) Immigrants are criminals simply by virtue of being immigrants \citep{ewing2015criminalization}. 

Figure \ref{fig:ideology} shows a clear ideological stratification for issue-specific frames: liberals favor \textit{hero} and \textit{victim} frames, while conservatives favor \textit{threat} frames. This finding is consistent with prior work on the role perceived threats play in shaping white American attitudes towards immigration ~\citep{Brader2008}, and the disposition of political conservatism to avoid potential threats \citep{jost2003political}. 

Second, while all frame categories show ideological bias, issue-specific frames are the most extreme.
Most notably, our analysis shows that focusing solely on issue-generic policy frames would obscure important patterns.
For example, the issue-generic \textit{cultural identity} frame shows a slight liberal bias; yet, related issue-specific frames diverge:  \textit{hero: cultural diversity} is very liberal while \textit{threat: national cohesion} is very conservative. 

Similarly, the  issue-generic \textit{economic} policy frame is slightly favored by more conservative authors, but the related \textit{issue-specific} frames \textit{threat: jobs} and \textit{hero: worker} reveal ideological divides.
This finding highlights the importance of using multiple framing typologies to provide a more nuanced analysis of immigration discourse.

Third, 
more liberal authors tend to use \textit{episodic} frames, while conservative authors tend to use \textit{thematic} frames. This difference is consistent with \citet{Somaini2019}'s finding that a local liberal newspaper featured more \textit{episodic} framing in immigration coverage, but a comparable conservative newspaper featured more \textit{thematic} framing. 
Other efforts that examine the relationship between narrative frames and cognitive and emotional responses provide some clues for the observed pattern. For instance, 
\citet{Aaroe2011} shows that thematic frames are stronger when there are no or weak emotional responses; and that the opposite is true for episodic frames. The divergence of findings could be driven by partisans' differing emotional responses.
Our findings also highlight important consequences for opinion formation. \citet{Iyengar1990} shows that episodic framing diverts attention from societal and political party responsibility; our results suggest that liberal Twitter users are likely to produce (and, due to partisan self-segregation, consume) social media content with such effects.
%
%


\begin{figure}[!h]
    \centering
    \includegraphics[height=5in]{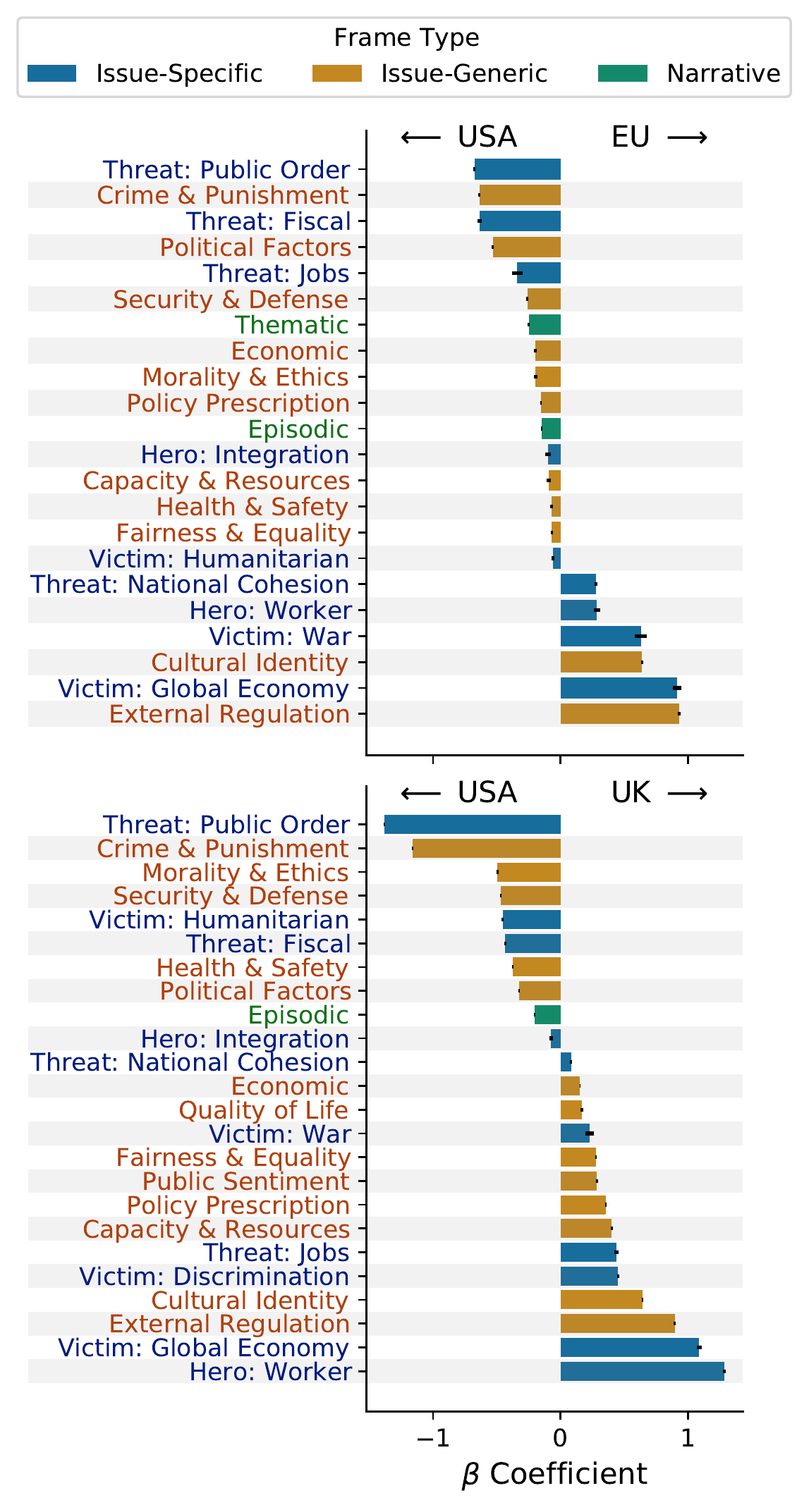}
    \caption{Effect of author being from the EU (top) or the UK (bottom) relative to the US.
Frames with positive $\beta$ coefficients are associated with authors from the EU (top) and UK (bottom), and frames with negative values are associated with US-based authors. Frames not significantly associated with region after Holm-Bonferroni correction are not included. 
}
\label{fig:country}
\end{figure}




\myparagraph{Region}
Immigration framing depends heavily on one's geopolitical entity (US, UK, and EU), as shown in Figure~\ref{fig:country}.
%
Several notable themes emerge.
First, many ideologically-extreme frames in the US, including \textit{crime \& punishment}, \textit{security \& defense}, \textit{threat: public order}, and \textit{threat: fiscal} are all significantly more likely to be found in US-based tweets relative to the UK and EU. This pattern suggests that region and ideology, and likely many other factors, interact in intricate ways to shape how ordinary people frame political issues. 

%

Second, \textit{cultural identity} is more strongly associated with both the UK and EU than the US. 
Perhaps immigrants' backgrounds are more marked in European discourse than in US discourse because the UK and EU have longer histories of cultural and ethnic homogeneity \cite{thorb2015}. This finding also reflects that Europeans' attitudes towards immigration depend on where immigrants are from and parallels how European newspapers frame immigration differently depending on migrants' countries of origin \citep{Eberl2018}.   

Finally, the bottom of Figure \ref{fig:country} shows that users from the UK are more likely to invoke labor-related frames. This prevalence of labor and economic frames has also been found in British traditional media \citep{Caviedes2015,Lawlor2015}, and has been attributed to differences in the labor market. Unlike migrants in the US, Italy, and France, who often work clandestinely in different economic sectors than domestic workers, UK migrants have proper authorization and are thus viewed as competition for British workers because they can work in the same industries \citep{Caviedes2015}.

\section{Audience Response to Frames}

\begin{figure}[!ht]
    \centering
    \includegraphics[width=\linewidth]{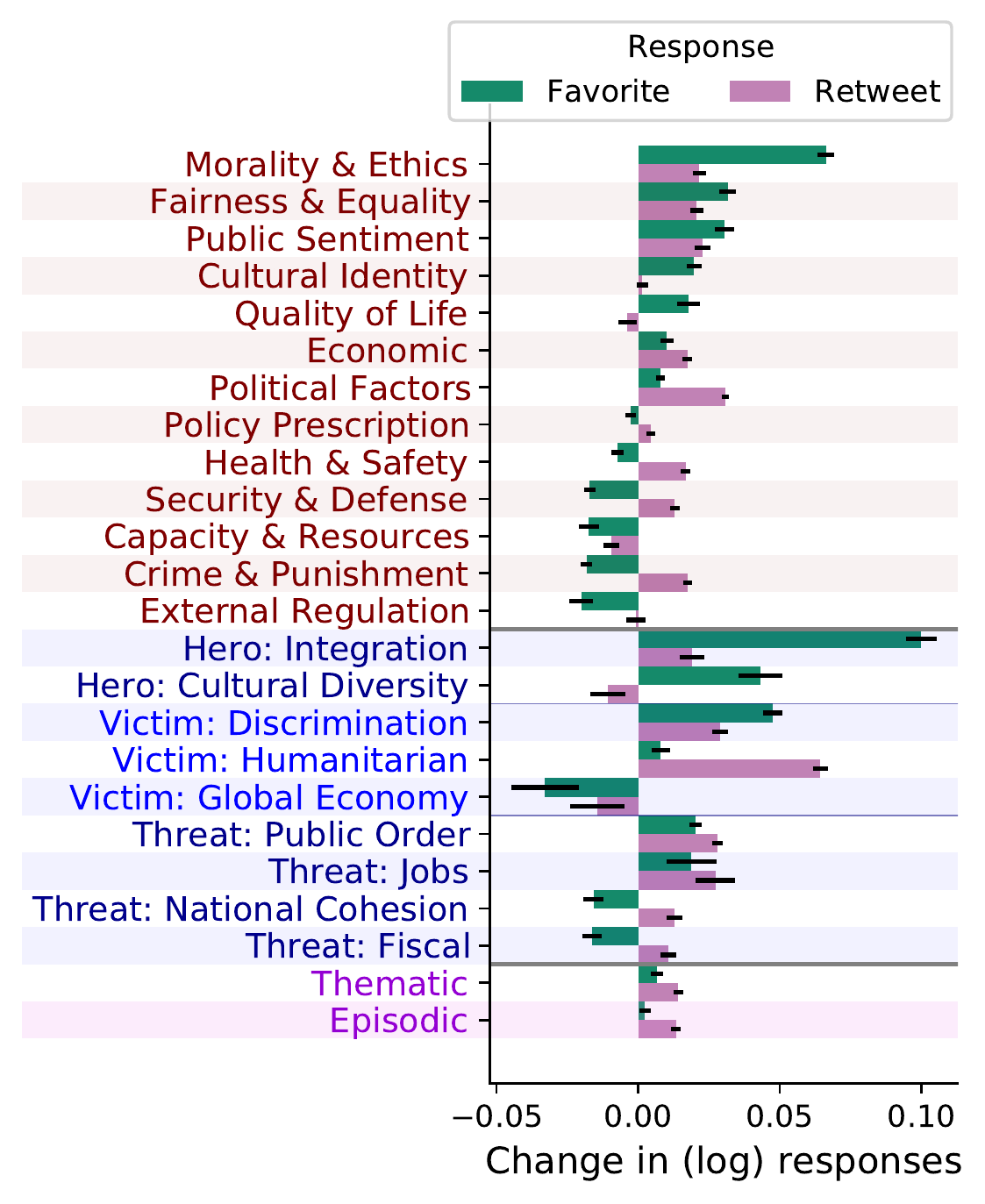}
    \caption{Effects of framing on two audience responses: favorites and retweets. The x-axis shows regression coefficients for the presence of each frame in predicting the log-scaled number of responses.
    Along the y-axis are all issue-generic policy frames (top), immigration-specific frames (middle), and narrative frames (bottom) that are significantly associated with either the number of favorites or retweets. 
    }
    \label{fig:responses}
\end{figure}

\citet[][p.~116]{Chong2007} assert that a ``challenge for future work concerns the identification of factors that make a frame strong.'' Studies of \textit{frame-setting}---i.e., how a message's framing affects its audience's emotions, beliefs, and opinions---have largely been restricted to small-scale experimental studies because responses to news media framing cannot be directly observed \citep{Eberl2018}. However, Twitter provides insight into the frame-setting process via interactive signals: favorites and retweets. While related, these two actions can have distinct underlying motivations: favoriting often indicates positive alignment between the author and the reader; in contrast, retweeting 
may also be driven by other motivations, such as the desire to inform or entertain others \citep{boyd2010}.
Different audience interactions have been shown to exhibit distinct patterns in political communication on Twitter \citep{minot2020ratioing}. 
Here, we test how a message's framing impacts both the favorites and retweets that it receives. 

\myparagraph{Experimental Setup}
We fit hierarchical linear mixed effects models with favorites and retweets (log-transformed) as the dependent variable on US tweets with detected author ideology. The presence of a frame is treated as a binary fixed effect. We control for all temporal, user-level and tweet-level features as in the prior section, as well as ideology. 

\myparagraph{Results}
The framing of immigration has a significant impact on how users engage with the content via retweets and favorites (Figure \ref{fig:responses}). 
Many issue-specific frames have a stronger effect on audience responses than either of the other typologies. As recent NLP approaches have adopted issue-generic frames for analysis \citep[e.g.,][]{kwak2020systematic},  the strength of issue-specific frames highlights the importance of expanding computational analyses beyond issue-generic frames, as other frames may have larger consequences for public opinion.

Most frames impact favorites and retweets differently, suggesting that the strength of a frame's effects is tied to the specific engagement behavior. Cultural frames (e.g. \textit{hero: integration}) and frames oriented around human interest (e.g. \textit{morality}, \textit{victim: discrimination}) are particularly associated with more endorsements (favorites), perhaps due to their increased emotional appeal to readers \cite{semetko2000framing}. 

On the other hand, \textit{political factors \& implications} is most highly associated with increased retweets. As the political frame emphasizes competition and strategy \cite{Boydstun2013}, this result mirrors similar links between the ``horse-race" frame in news reports and engagement \citep{iyengar2004consumer}; users may prefer amplifying political messages via retweeting to help their side win. 

Similarly, frames about security and safety (e.g. \textit{crime \& punishment}, \textit{victim: humanitarian}) are highly associated with more retweets, but not necessarily favorites. While security and safety frames may not lead audience members to endorse such messages, perhaps they are more likely to amplify these messages due to perceived urgency or the desire to persuade others of such concerns.

Finally, Figure \ref{fig:responses} shows how a message's narrative framing impacts audience response, even after controlling for all other frames. Both \textit{episodic} and \textit{thematic} frames are significantly associated with increased engagement (retweets), but less strongly than issue frames. Having a clear narrative is important for messages to spread, but the underlying mechanisms driving engagement behaviors may differ for \textit{episodic} and \textit{thematic} frames; prior work on mainstream media has found that news stories using episodic frames tend to be more emotionally engaging, while thematic frames can be more persuasive \citep{iyengar1991anyone,gross2008}.

\section{Conclusion} 
Users' exposure to political information on social media can have immense consequences. By leveraging multiple theory-informed typologies, our computational analysis of framing enables us to better understand public discourses surrounding immigration. We furthermore show that framing on Twitter affects how audience interactions with messages via favoriting and retweeting behaviors. This work has implications for social media platforms, who may wish to improve users' experiences by enabling them to discover content with a diversity of frames. By exposing users to a wide range of perspectives, this work can help lay foundations for more cooperative and effective online discussions. All code, data, annotation guidelines, and pretrained models are available at {\small \url{https://github.com/juliamendelsohn/framing}}.

\section{Ethical Considerations}

Our analysis of frame-building involves inferring political ideology and regional from users with existing tools, so we aggregated this information in our analysis in order to minimize the risk of exposing potentially sensitive personal data about individuals. Our dataset includes tweet IDs along with frame labels, but no additional social information. However, there are also ethical consequences of categorizing people along these social dimensions. We acknowledge that reducing people's social identities to region and ideology obscures the wide range of unobservable and non-quantifiable predispositions and experiences that may impact framing and attitudes towards immigration. 

We emphasize that our dataset is not fully representative of all immigration discourse and should not be treated as such. Twitter's demographics are not representative of the global population \citep{mislove2011understanding}. Furthermore, our dataset only includes tweets with authors from particular Western countries. All tweets were automatically identified by Twitter as being written in English, thus additionally imposing standard language ideologies on the data that we include \citep{milroy2001language}. Furthermore, language choice itself can be a socially and politically meaningful linguistic cue that may have unique interactions with framing \citep[e.g.,][]{gal1978peasant,shoemark2017aye,stewart2018si,ndubuisi2019wetin}. 

Although we do not focus on abusive language, our topical content contains frequent instances of racism, Islamophobia, antisemitism, and personal insults. We caution future researchers about potentially traumatic psychological effects of working with this dataset.

We aim to support immigrants, an often marginalized group, by shedding light on their representation on social media. However, there is a risk that malicious agents could exploit our frame-setting findings by disseminating harmful content packaged in more popular frames.

\section{Acknowledgements}

We thank Anoop Kotha, Shiqi Sheng, Guoxin Yin, and Hongting Zhu for their contributions to the data annotation effort. We also thank Libby Hemphill and Stuart Soroka for their valuable comments and feedback. This work was supported in part through funding from the Volkswagen Foundation.

\bibliography{arxiv}
\bibliographystyle{acl_natbib}

\clearpage
\newpage

\appendix

\section{Frame distribution in annotated data}
\begin{figure}[h]
    \centering
    \includegraphics[width=\linewidth]{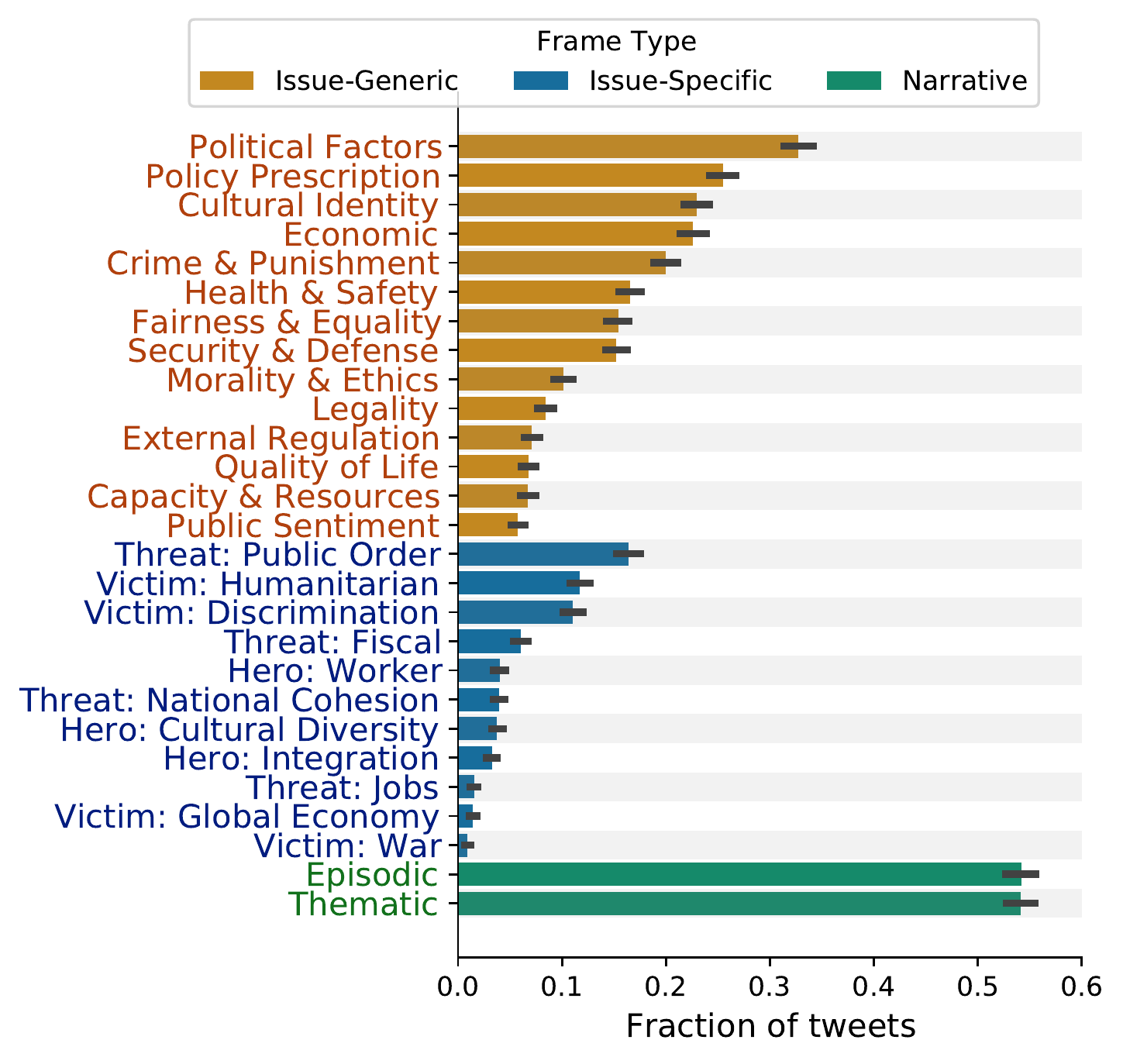}
    \vspace{-2em}
    \caption{Distribution of frames in annotated data.}
    \label{fig:distribution}
\end{figure}

Figure \ref{fig:distribution} shows the distribution of frames as a fraction of total tweets in the annotated data.
\label{app:distributions}

\vspace{-2em}
\section{Inter-annotator agreement plots} 
\begin{figure}[!h]
    \centering
    \includegraphics[height=2in,width=\linewidth]{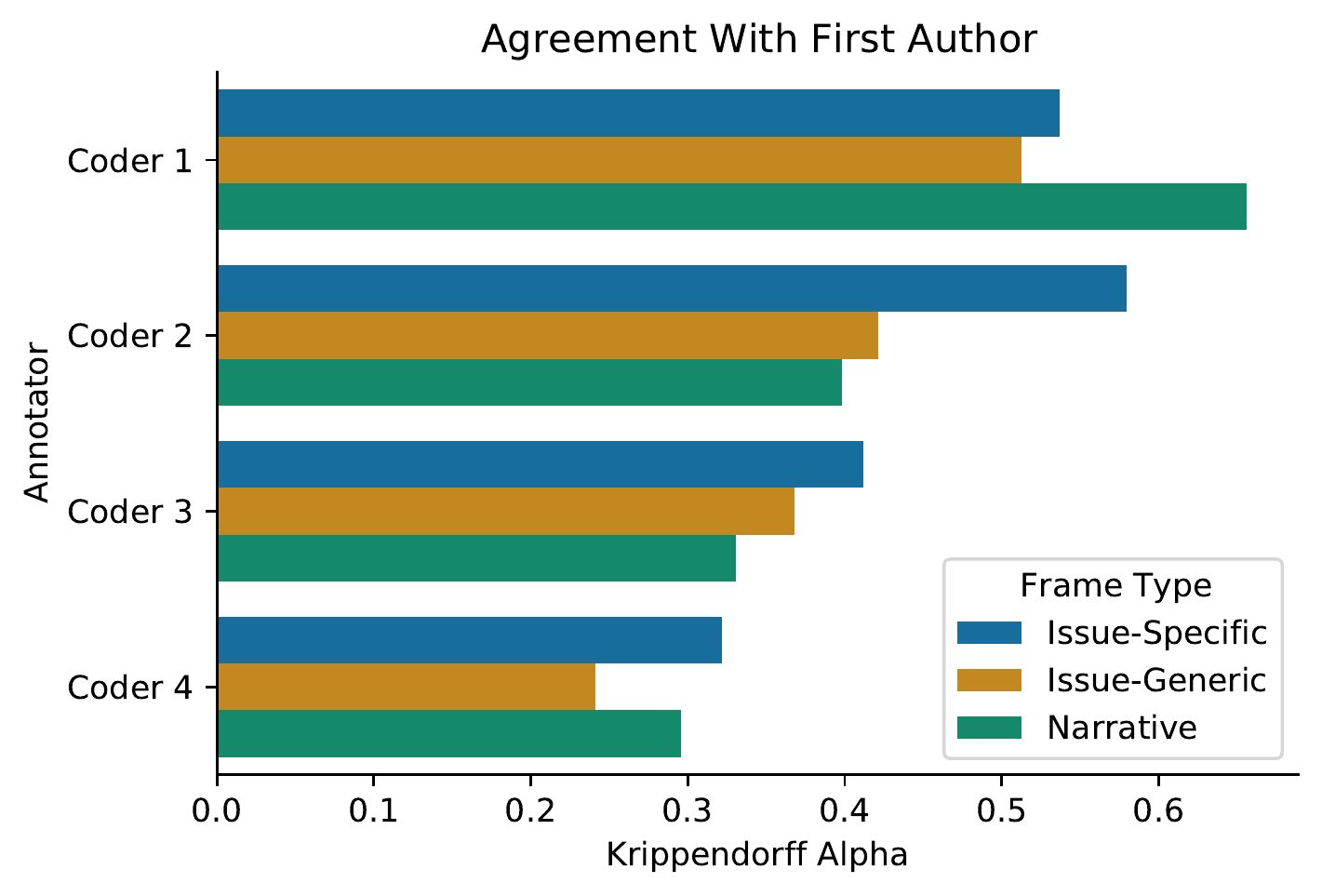}
    \vspace{-2em}
    \caption{Inter-annotator agreement between first author and other coders before consensus-coding.}
    \label{fig:agreement_julia}
\end{figure}

\begin{figure}[!h]
    \centering
    \includegraphics[height=2in,width=\linewidth]{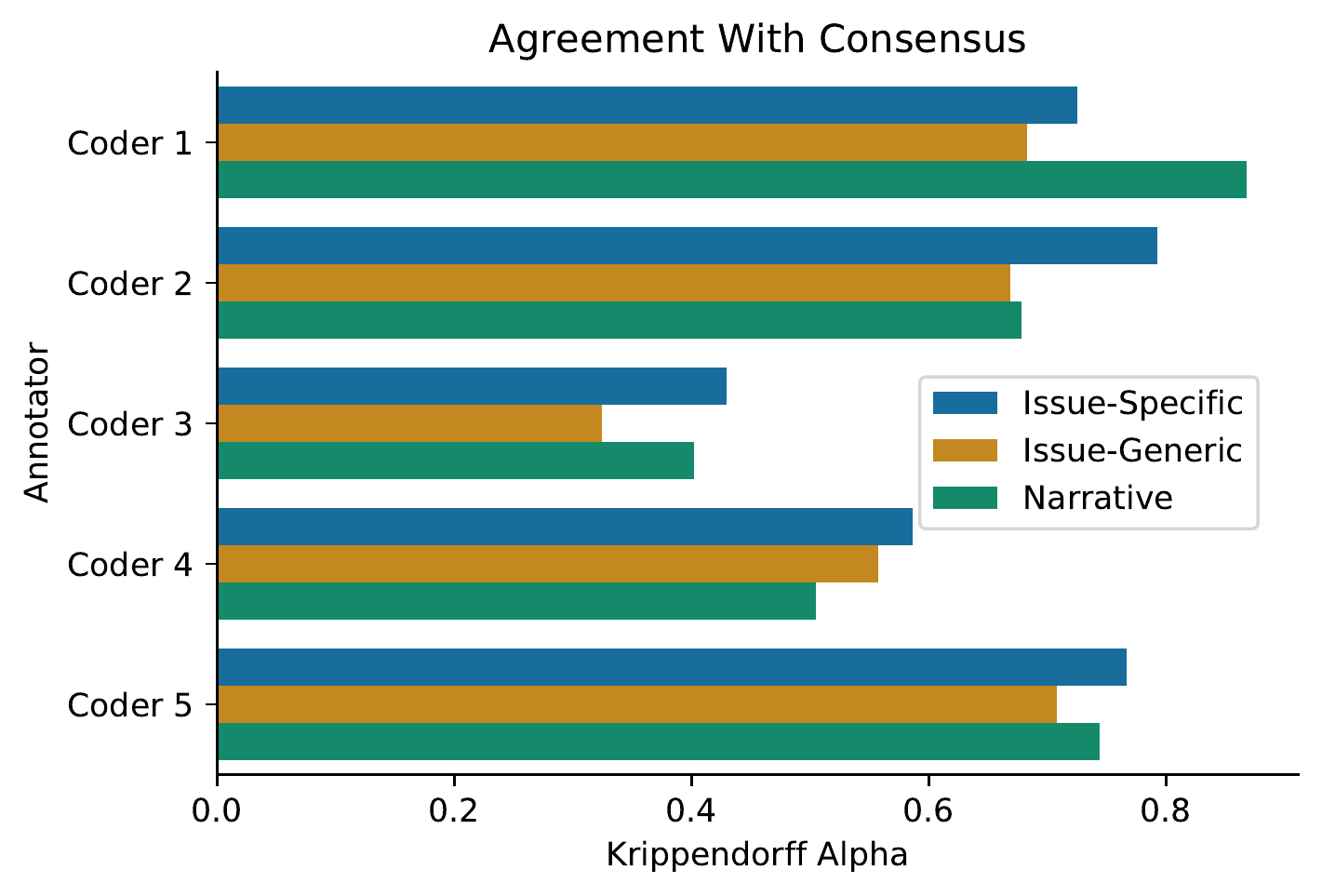}
    \vspace{-2em}
    \caption{Agreement between each coder and consensus annotations before consensus-coding.}
    \label{fig:agreement_consensus}
\end{figure}

Figures \ref{fig:agreement_julia} and \ref{fig:agreement_consensus} show inter-annotator agreement (Krippendorff's $\alpha$) across frame types. 

\label{app:agreement}

 \vspace{-2em}

\newpage
\section{Frame detection performance} 
\label{app:model-performance}
\begin{figure}[!htbp]
    \centering
    \includegraphics[height=2in,width=\linewidth]{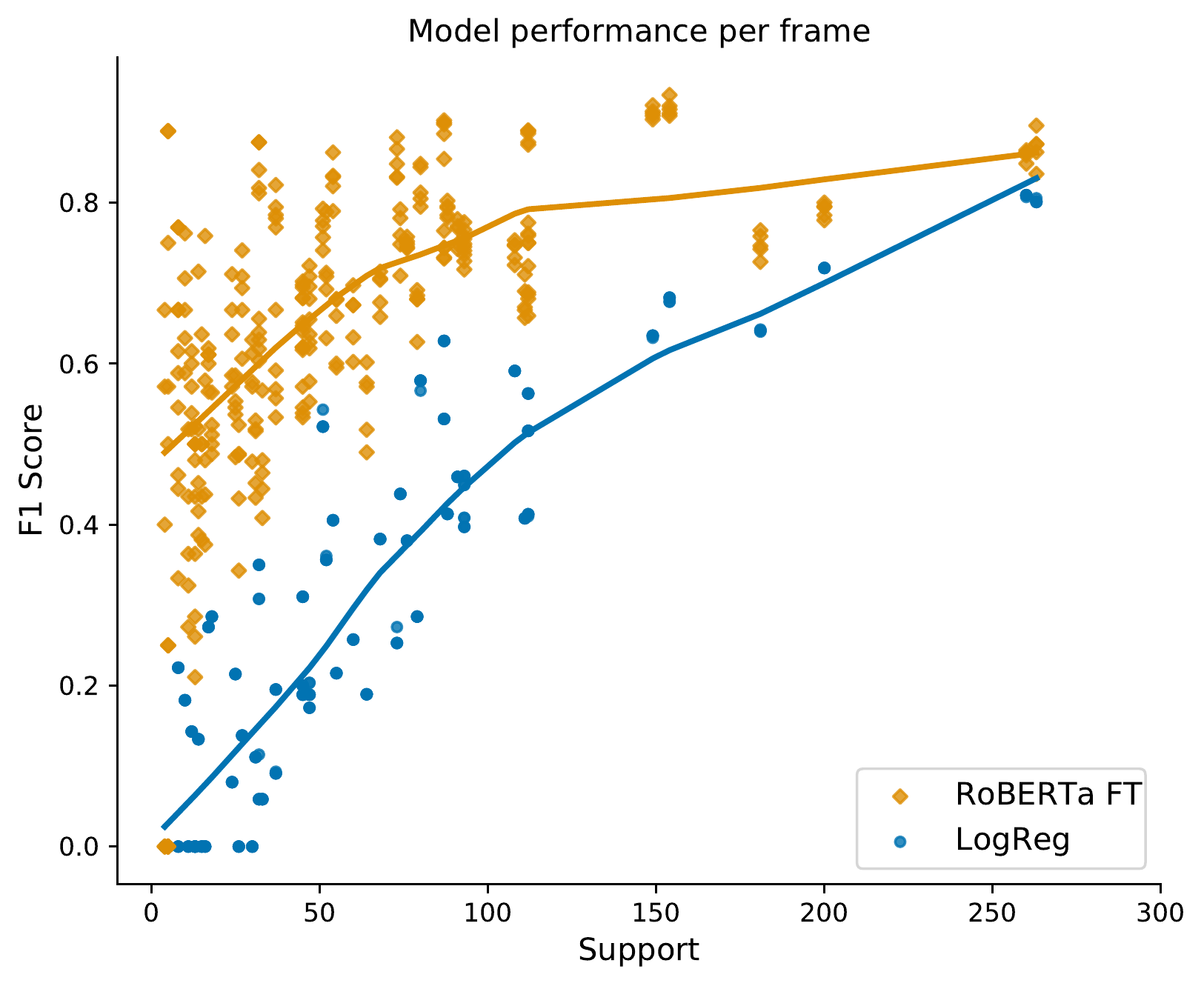}
    \vspace{-2em}

    \caption{F1 score of logistic regression (1,2-gram features) and fine-tuned RoBERTa for each frame and frame support in evaluation sets. RoBERTa consistently outperforms logistic regression, especially for low-frequency frames. 
    }
    \label{fig:support_f1}
\end{figure}

\begin{figure}[!htbp]
    \centering

    \includegraphics[height=2in,width=\linewidth]{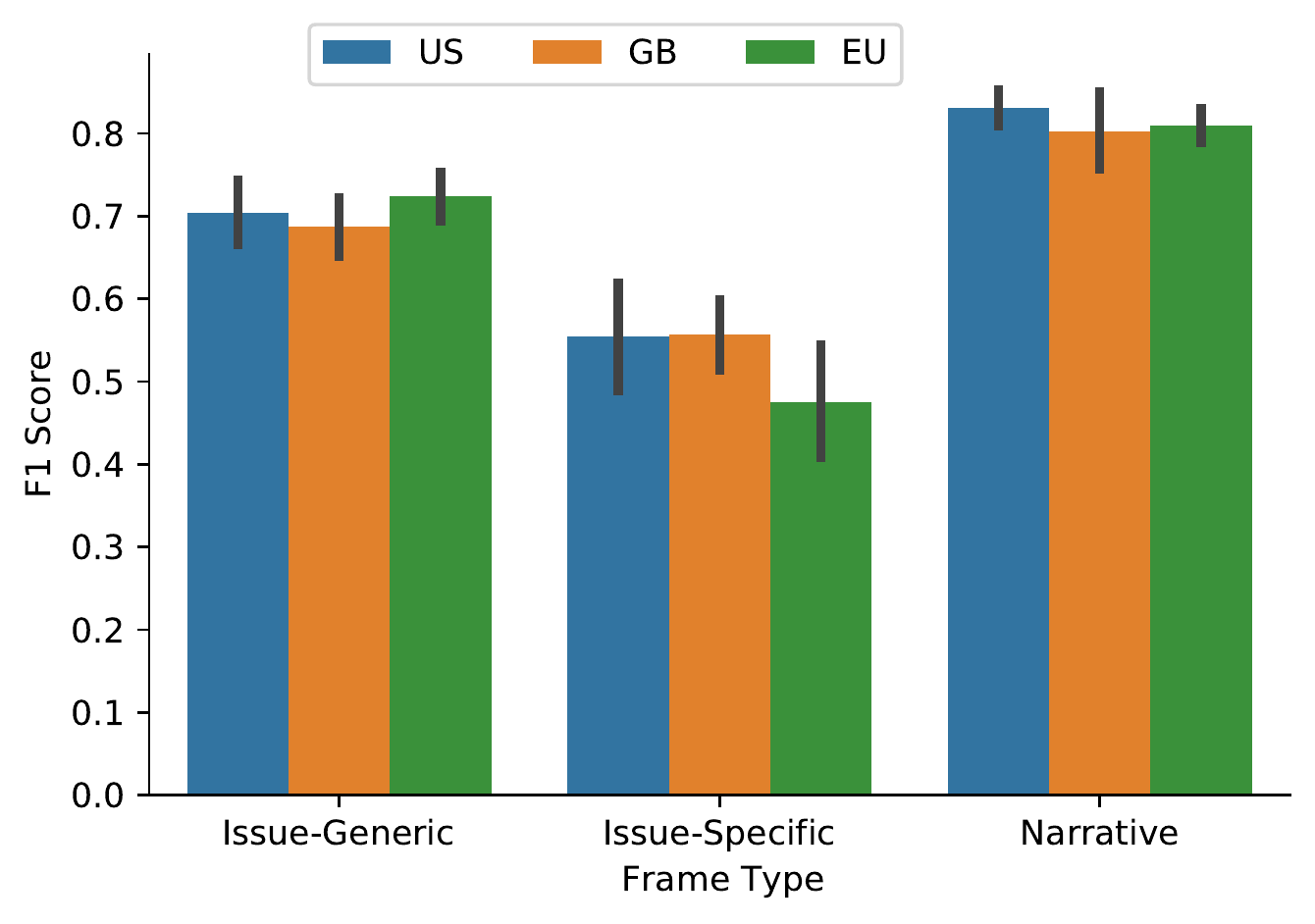}
    \vspace{-2em}

    \caption{Average F1 scores on combined dev/test set separated by region. Models achieve comparable performance for the United States, United Kingdom, and European Union, except for slightly lower performance for issue-specific frames on EU tweets.}
    \label{fig:country_f1}
\end{figure}

Tables \ref{tab:dev_by_type}-\ref{tab:test_by_frame}
 and Figures \ref{fig:support_f1}-\ref{fig:country_f1} provide details about the fine-tuned RoBERTa models' performance.

\begin{table}[!h]
\centering

\resizebox{\linewidth}{!}{%
\begin{tabular}{@{}ccccc@{}}
\textbf{Frame Type} & \textbf{Precision} & \textbf{Recall} & \textbf{F1-score} & \textbf{LRAP} \\ \hline
Issue-Generic Policy & 0.722 & 0.727 & 0.716 & 0.745 \\
Issue-Specific & 0.667 & 0.493 & 0.550 & 0.785 \\
Narrative & 0.780 & 0.884 & 0.825 & 0.896
\end{tabular}%
}

\caption{Performance by frame type on dev set.}
\label{tab:dev_by_type}
\end{table}

\begin{table}[!h]
\centering
\vspace{-1em}

\resizebox{\linewidth}{!}{%
\begin{tabular}{cccc}
 & \textbf{Issue-Generic} & \textbf{Issue-Specific} & \textbf{Narrative} \\ \hline
Human-Machine & 0.443 & 0.488 & 0.421 \\
Human-Human & 0.417 & 0.491 & 0.458 \\ 
\end{tabular}%
}
\caption{Average Krippendorff $\alpha$ agreement between human annotators and machine-predicted labels (top row) and between human annotator pairs (bottom row). Overall, our classifiers had similar agreement with human annotators as humans did with one another. }
\label{tab:human-machine-agreement}
\end{table}

\begin{table*}[!htbp]
\centering
\resizebox{.85\textwidth}{!}{%
\begin{tabular}{@{}ccccccc@{}}
\textbf{Frame Type} & \textbf{Frame} & \textbf{Precision} & \textbf{Recall} & \textbf{F1-score} & \textbf{Support} & \textbf{LRAP} \\
\multirow{14}{*}{\textbf{Issue-General}} & Capacity and Resources & 0.530 & 0.706 & 0.601 & 17.0 & 0.745 \\
 & Crime and Punishment & 0.787 & 0.798 & 0.791 & 88.0 & 0.745 \\
 & Cultural Identity & 0.716 & 0.815 & 0.762 & 91.0 & 0.745 \\
 & Economic & 0.821 & 0.968 & 0.888 & 87.0 & 0.745 \\
 & External Regulation and Reputation & 0.658 & 0.493 & 0.562 & 45.0 & 0.745 \\
 & Fairness and Equality & 0.678 & 0.712 & 0.692 & 68.0 & 0.745 \\
 & Health and Safety & 0.841 & 0.866 & 0.852 & 73.0 & 0.745 \\
 & Legality, Constitutionality, Jurisdiction & 0.764 & 0.822 & 0.790 & 37.0 & 0.745 \\
 & Morality and Ethics & 0.792 & 0.549 & 0.643 & 55.0 & 0.745 \\
 & Policy Prescription and Evaluation & 0.689 & 0.807 & 0.740 & 108.0 & 0.745 \\
 & Political Factors and Implications & 0.890 & 0.948 & 0.918 & 154.0 & 0.745 \\
 & Public Sentiment & 0.619 & 0.388 & 0.473 & 33.0 & 0.745 \\
 & Quality of Life & 0.502 & 0.484 & 0.490 & 31.0 & 0.745 \\
 & Security and Defense & 0.825 & 0.818 & 0.821 & 80.0 & 0.745 \\
\multirow{11}{*}{\textbf{Issue-Specific}} & Hero: Cultural Diversity & 0.699 & 0.387 & 0.490 & 15.0 & 0.785 \\
 & Hero: Integration & 0.455 & 0.354 & 0.391 & 13.0 & 0.785 \\
 & Hero: Worker & 0.501 & 0.369 & 0.421 & 13.0 & 0.785 \\
 & Threat: Fiscal & 0.613 & 0.504 & 0.541 & 25.0 & 0.785 \\
 & Threat: Jobs & 1.000 & 0.525 & 0.683 & 8.0 & 0.785 \\
 & Threat: National Cohesion & 0.565 & 0.512 & 0.526 & 16.0 & 0.785 \\
 & Threat: Public Order & 0.773 & 0.746 & 0.758 & 74.0 & 0.785 \\
 & Victim: Discrimination & 0.615 & 0.684 & 0.646 & 45.0 & 0.785 \\
 & Victim: Global Economy & 0.933 & 0.600 & 0.720 & 5.0 & 0.785 \\
 & Victim: Humanitarian & 0.647 & 0.488 & 0.551 & 64.0 & 0.785 \\
 & Victim: War & 0.533 & 0.250 & 0.328 & 4.0 & 0.785 \\
\multirow{2}{*}{\textbf{Narrative}} & Episodic & 0.706 & 0.902 & 0.791 & 200.0 & 0.896 \\
 & Thematic & 0.855 & 0.865 & 0.859 & 260.0 & 0.896
\end{tabular}%
}
\caption{Performance per frame on development set}
\label{tab:dev_by_frame}
\end{table*}

\begin{table*}[!htbp]
\centering
\resizebox{.85\textwidth}{!}{%
\begin{tabular}{@{}ccccccc@{}}
\textbf{Frame Type} & \textbf{Frame} & \textbf{Precision} & \textbf{Recall} & \textbf{F1-score} & \textbf{Support} & \textbf{LRAP} \\
\multirow{14}{*}{\textbf{Issue-General}} & Capacity and Resources & 0.451 & 0.611 & 0.517 & 18.0 & 0.750 \\
 & Crime and Punishment & 0.817 & 0.695 & 0.749 & 76.0 & 0.750 \\
 & Cultural Identity & 0.687 & 0.852 & 0.760 & 93.0 & 0.750 \\
 & Economic & 0.824 & 0.950 & 0.882 & 112.0 & 0.750 \\
 & External Regulation and Reputation & 0.708 & 0.581 & 0.629 & 32.0 & 0.750 \\
 & Fairness and Equality & 0.721 & 0.635 & 0.673 & 79.0 & 0.750 \\
 & Health and Safety & 0.784 & 0.878 & 0.828 & 54.0 & 0.750 \\
 & Legality, Constitutionality, Jurisdiction & 0.817 & 0.875 & 0.844 & 32.0 & 0.750 \\
 & Morality and Ethics & 0.698 & 0.570 & 0.623 & 47.0 & 0.750 \\
 & Policy Prescription and Evaluation & 0.660 & 0.855 & 0.743 & 87.0 & 0.750 \\
 & Political Factors and Implications & 0.912 & 0.911 & 0.911 & 149.0 & 0.750 \\
 & Public Sentiment & 0.713 & 0.338 & 0.455 & 26.0 & 0.750 \\
 & Quality of Life & 0.657 & 0.520 & 0.574 & 30.0 & 0.750 \\
 & Security and Defense & 0.725 & 0.816 & 0.768 & 51.0 & 0.750 \\ 
\multirow{11}{*}{\textbf{Issue-Specific}} & Hero: Cultural Diversity & 0.591 & 0.567 & 0.569 & 12.0 & 0.806 \\
 & Hero: Integration & 0.503 & 0.500 & 0.498 & 14.0 & 0.806 \\
 & Hero: Worker & 0.710 & 0.575 & 0.634 & 24.0 & 0.806 \\
 & Threat: Fiscal & 0.694 & 0.689 & 0.683 & 27.0 & 0.806 \\
 & Threat: Jobs & 0.743 & 0.620 & 0.671 & 10.0 & 0.806 \\
 & Threat: National Cohesion & 0.344 & 0.455 & 0.383 & 11.0 & 0.806 \\
 & Threat: Public Order & 0.737 & 0.681 & 0.707 & 52.0 & 0.806 \\
 & Victim: Discrimination & 0.785 & 0.570 & 0.656 & 60.0 & 0.806 \\
 & Victim: Global Economy & 0.571 & 0.450 & 0.489 & 8.0 & 0.806 \\
 & Victim: Humanitarian & 0.715 & 0.658 & 0.681 & 45.0 & 0.806 \\
 & Victim: War & 0.133 & 0.080 & 0.100 & 5.0 & 0.806 \\
\multirow{2}{*}{\textbf{Narrative}} & Episodic & 0.630 & 0.922 & 0.748 & 181.0 & 0.894 \\
 & Thematic & 0.885 & 0.852 & 0.868 & 263.0 & 0.894
\end{tabular}%
}
\caption{Performance per frame on test set}
\label{tab:test_by_frame}
\end{table*}

\newpage

\section{Frame-building (region) regressions}
\label{app:country-regressions}

Tables \ref{tab:reg_country_start}-\ref{tab:reg_country_end} show independent variable coefficients for logit regressions predicting frames from region.

\begin{table}[!htbp] \centering 
  \caption{Capacity.and.Resources} 
  \vspace{-1em}

  \label{tab:reg_country_start} 
\resizebox{.8\columnwidth}{!}{%
\begin{tabular}{@{\extracolsep{5pt}}lc} 

\hline \\[-1.8ex] 
 countryEU & $-$0.091$^{***}$ (0.015) \\ 
  countryGB & 0.403$^{***}$ (0.007) \\ 
  has\_hashtag1 & $-$0.301$^{***}$ (0.011) \\ 
  has\_mention1 & 0.057$^{***}$ (0.009) \\ 
  has\_url1 & 0.001 (0.011) \\ 
  is\_quote\_status1 & 0.324$^{***}$ (0.009) \\ 
  is\_reply1 & 0.359$^{***}$ (0.009) \\ 
  is\_verified1 & $-$0.239$^{***}$ (0.020) \\ 
  log\_chars & 0.730$^{***}$ (0.008) \\ 
  log\_followers & $-$0.019$^{***}$ (0.003) \\ 
  log\_following & 0.006$^{**}$ (0.003) \\ 
  log\_statuses & $-$0.005$^{**}$ (0.002) \\ 
  Constant & $-$6.801$^{***}$ (0.082) \\ 
 \hline \\[-1.8ex] 
Observations & 2,661,435 \\ 
Log Likelihood & $-$611,982.800 \\ 
Akaike Inf. Crit. & 1,223,998.000 \\ 
Bayesian Inf. Crit. & 1,224,202.000 \\ 
\hline 
\hline \\[-1.8ex] 
\textit{Note:}  & \multicolumn{1}{r}{$^{*}$p$<$0.1; $^{**}$p$<$0.05; $^{***}$p$<$0.01} \\ 
\end{tabular} }
\end{table} 

\vspace{-1em}
\begin{table}[!htbp] \centering 
  \caption{Crime.and.Punishment} 
  \vspace{-1em}

  \label{} 
 \resizebox{.8\columnwidth}{!}{%

\begin{tabular}{@{\extracolsep{5pt}}lc} 
\hline \\[-1.8ex] 
 countryEU & $-$0.635$^{***}$ (0.008) \\ 
  countryGB & $-$1.162$^{***}$ (0.006) \\ 
  has\_hashtag1 & $-$0.123$^{***}$ (0.005) \\ 
  has\_mention1 & $-$0.032$^{***}$ (0.004) \\ 
  has\_url1 & 0.175$^{***}$ (0.005) \\ 
  is\_quote\_status1 & 0.084$^{***}$ (0.005) \\ 
  is\_reply1 & 0.072$^{***}$ (0.005) \\ 
  is\_verified1 & $-$0.311$^{***}$ (0.009) \\ 
  log\_chars & 0.261$^{***}$ (0.004) \\ 
  log\_followers & $-$0.013$^{***}$ (0.001) \\ 
  log\_following & 0.031$^{***}$ (0.001) \\ 
  log\_statuses & 0.023$^{***}$ (0.001) \\ 
  Constant & $-$2.550$^{***}$ (0.038) \\ 
 \hline \\[-1.8ex] 
Observations & 2,661,435 \\ 
Log Likelihood & $-$1,598,017.000 \\ 
Akaike Inf. Crit. & 3,196,066.000 \\ 
Bayesian Inf. Crit. & 3,196,271.000 \\ 
\hline 
\hline \\[-1.8ex] 
\textit{Note:}  & \multicolumn{1}{r}{$^{*}$p$<$0.1; $^{**}$p$<$0.05; $^{***}$p$<$0.01} \\ 
\end{tabular} }
\end{table} 

\vspace{-1em}
\begin{table}[!htbp] \centering 
  \caption{Cultural.Identity} 
  \vspace{-1em}

  \label{} 
\resizebox{.8\columnwidth}{!}{%

\begin{tabular}{@{\extracolsep{5pt}}lc} 
\hline \\[-1.8ex] 
 countryEU & 0.641$^{***}$ (0.008) \\ 
  countryGB & 0.643$^{***}$ (0.005) \\ 
  has\_hashtag1 & $-$0.102$^{***}$ (0.006) \\ 
  has\_mention1 & $-$0.357$^{***}$ (0.006) \\ 
  has\_url1 & $-$0.322$^{***}$ (0.007) \\ 
  is\_quote\_status1 & 0.008 (0.006) \\ 
  is\_reply1 & 0.228$^{***}$ (0.006) \\ 
  is\_verified1 & $-$0.065$^{***}$ (0.011) \\ 
  log\_chars & 0.758$^{***}$ (0.005) \\ 
  log\_followers & 0.029$^{***}$ (0.002) \\ 
  log\_following & $-$0.043$^{***}$ (0.002) \\ 
  log\_statuses & $-$0.022$^{***}$ (0.001) \\ 
  Constant & $-$4.890$^{***}$ (0.043) \\ 
 \hline \\[-1.8ex] 
Observations & 2,661,435 \\ 
Log Likelihood & $-$1,162,059.000 \\ 
Akaike Inf. Crit. & 2,324,150.000 \\ 
Bayesian Inf. Crit. & 2,324,354.000 \\ 
\hline 
\hline \\[-1.8ex] 
\textit{Note:}  & \multicolumn{1}{r}{$^{*}$p$<$0.1; $^{**}$p$<$0.05; $^{***}$p$<$0.01} \\ 
\end{tabular} }
\end{table} 

\begin{table}[!htbp] \centering 
  \caption{Economic} 
  \label{} 
  
\resizebox{.8\columnwidth}{!}{%

\begin{tabular}{@{\extracolsep{5pt}}lc} 

\hline \\[-1.8ex] 
 countryEU & $-$0.197$^{***}$ (0.009) \\ 
  countryGB & 0.154$^{***}$ (0.005) \\ 
  has\_hashtag1 & $-$0.129$^{***}$ (0.006) \\ 
  has\_mention1 & $-$0.010$^{**}$ (0.005) \\ 
  has\_url1 & 0.017$^{***}$ (0.006) \\ 
  is\_quote\_status1 & 0.208$^{***}$ (0.005) \\ 
  is\_reply1 & 0.199$^{***}$ (0.005) \\ 
  is\_verified1 & $-$0.089$^{***}$ (0.010) \\ 
  log\_chars & 0.916$^{***}$ (0.005) \\ 
  log\_followers & $-$0.053$^{***}$ (0.001) \\ 
  log\_following & 0.027$^{***}$ (0.002) \\ 
  log\_statuses & 0.031$^{***}$ (0.001) \\ 
  Constant & $-$6.173$^{***}$ (0.052) \\ 
 \hline \\[-1.8ex] 
Observations & 2,661,435 \\ 
Log Likelihood & $-$1,409,032.000 \\ 
Akaike Inf. Crit. & 2,818,096.000 \\ 
Bayesian Inf. Crit. & 2,818,301.000 \\ 
\hline 
\hline \\[-1.8ex] 
\textit{Note:}  & \multicolumn{1}{r}{$^{*}$p$<$0.1; $^{**}$p$<$0.05; $^{***}$p$<$0.01} \\ 
\end{tabular} }
\end{table} 


\begin{table}[!htbp] \centering 
  \caption{External.Regulation.and.Reputation} 
  \label{} 
\resizebox{.8\columnwidth}{!}{%

\begin{tabular}{@{\extracolsep{5pt}}lc} 
\hline \\[-1.8ex] 
 countryEU & 0.934$^{***}$ (0.012) \\ 
  countryGB & 0.898$^{***}$ (0.008) \\ 
  has\_hashtag1 & 0.112$^{***}$ (0.011) \\ 
  has\_mention1 & 0.016 (0.010) \\ 
  has\_url1 & 0.673$^{***}$ (0.014) \\ 
  is\_quote\_status1 & $-$0.044$^{***}$ (0.011) \\ 
  is\_reply1 & $-$0.058$^{***}$ (0.010) \\ 
  is\_verified1 & 0.020 (0.020) \\ 
  log\_chars & 0.551$^{***}$ (0.009) \\ 
  log\_followers & $-$0.014$^{***}$ (0.003) \\ 
  log\_following & $-$0.011$^{***}$ (0.003) \\ 
  log\_statuses & 0.047$^{***}$ (0.002) \\ 
  Constant & $-$7.119$^{***}$ (0.087) \\ 
 \hline \\[-1.8ex] 
Observations & 2,661,435 \\ 
Log Likelihood & $-$448,512.800 \\ 
Akaike Inf. Crit. & 897,057.500 \\ 
Bayesian Inf. Crit. & 897,262.200 \\ 
\hline 
\hline \\[-1.8ex] 
\textit{Note:}  & \multicolumn{1}{r}{$^{*}$p$<$0.1; $^{**}$p$<$0.05; $^{***}$p$<$0.01} \\ 
\end{tabular} }
\end{table} 

\begin{table}[!htbp] \centering 
  \caption{Fairness.and.Equality} 
  \label{} 
\resizebox{.8\columnwidth}{!}{%

\begin{tabular}{@{\extracolsep{5pt}}lc} 
\hline \\[-1.8ex] 
 countryEU & $-$0.068$^{***}$ (0.010) \\ 
  countryGB & 0.280$^{***}$ (0.006) \\ 
  has\_hashtag1 & 0.029$^{***}$ (0.007) \\ 
  has\_mention1 & $-$0.243$^{***}$ (0.006) \\ 
  has\_url1 & $-$0.117$^{***}$ (0.008) \\ 
  is\_quote\_status1 & 0.511$^{***}$ (0.006) \\ 
  is\_reply1 & 0.566$^{***}$ (0.007) \\ 
  is\_verified1 & $-$0.031$^{**}$ (0.013) \\ 
  log\_chars & 0.933$^{***}$ (0.006) \\ 
  log\_followers & $-$0.034$^{***}$ (0.002) \\ 
  log\_following & 0.017$^{***}$ (0.002) \\ 
  log\_statuses & 0.002 (0.001) \\ 
  Constant & $-$6.774$^{***}$ (0.042) \\ 
 \hline \\[-1.8ex] 
Observations & 2,661,435 \\ 
Log Likelihood & $-$1,027,084.000 \\ 
Akaike Inf. Crit. & 2,054,199.000 \\ 
Bayesian Inf. Crit. & 2,054,404.000 \\ 
\hline 
\hline \\[-1.8ex] 
\textit{Note:}  & \multicolumn{1}{r}{$^{*}$p$<$0.1; $^{**}$p$<$0.05; $^{***}$p$<$0.01} \\ 
\end{tabular} }
\end{table} 

\begin{table}[!htbp] \centering 
  \caption{Health.and.Safety} 
  \label{} 
 \resizebox{.8\columnwidth}{!}{%

\begin{tabular}{@{\extracolsep{5pt}}lc} 
\hline \\[-1.8ex] 
 countryEU & $-$0.070$^{***}$ (0.009) \\ 
  countryGB & $-$0.375$^{***}$ (0.006) \\ 
  has\_hashtag1 & $-$0.061$^{***}$ (0.006) \\ 
  has\_mention1 & $-$0.134$^{***}$ (0.005) \\ 
  has\_url1 & 0.512$^{***}$ (0.007) \\ 
  is\_quote\_status1 & $-$0.148$^{***}$ (0.006) \\ 
  is\_reply1 & $-$0.140$^{***}$ (0.005) \\ 
  is\_verified1 & $-$0.072$^{***}$ (0.011) \\ 
  log\_chars & 0.432$^{***}$ (0.005) \\ 
  log\_followers & $-$0.026$^{***}$ (0.002) \\ 
  log\_following & 0.036$^{***}$ (0.002) \\ 
  log\_statuses & 0.028$^{***}$ (0.001) \\ 
  Constant & $-$4.279$^{***}$ (0.078) \\ 
 \hline \\[-1.8ex] 
Observations & 2,661,435 \\ 
Log Likelihood & $-$1,256,905.000 \\ 
Akaike Inf. Crit. & 2,513,841.000 \\ 
Bayesian Inf. Crit. & 2,514,046.000 \\ 
\hline 
\hline \\[-1.8ex] 
\textit{Note:}  & \multicolumn{1}{r}{$^{*}$p$<$0.1; $^{**}$p$<$0.05; $^{***}$p$<$0.01} \\ 
\end{tabular} }
\end{table}

\begin{table}[!htbp] \centering 
  \caption{Legality..Constitutionality..Jurisdiction} 
  \label{} 
\resizebox{.8\columnwidth}{!}{%
\begin{tabular}{@{\extracolsep{5pt}}lc} 

\hline \\[-1.8ex] 
 countryEU & $-$0.494$^{***}$ (0.012) \\ 
  countryGB & $-$0.688$^{***}$ (0.008) \\ 
  has\_hashtag1 & $-$0.041$^{***}$ (0.007) \\ 
  has\_mention1 & $-$0.012$^{**}$ (0.006) \\ 
  has\_url1 & 0.263$^{***}$ (0.008) \\ 
  is\_quote\_status1 & $-$0.206$^{***}$ (0.007) \\ 
  is\_reply1 & $-$0.200$^{***}$ (0.006) \\ 
  is\_verified1 & 0.363$^{***}$ (0.012) \\ 
  log\_chars & 0.734$^{***}$ (0.006) \\ 
  log\_followers & $-$0.050$^{***}$ (0.002) \\ 
  log\_following & 0.024$^{***}$ (0.002) \\ 
  log\_statuses & 0.058$^{***}$ (0.001) \\ 
  Constant & $-$6.193$^{***}$ (0.041) \\ 
 \hline \\[-1.8ex] 
Observations & 2,661,435 \\ 
Log Likelihood & $-$968,625.200 \\ 
Akaike Inf. Crit. & 1,937,282.000 \\ 
Bayesian Inf. Crit. & 1,937,487.000 \\ 
\hline 
\hline \\[-1.8ex] 
\textit{Note:}  & \multicolumn{1}{r}{$^{*}$p$<$0.1; $^{**}$p$<$0.05; $^{***}$p$<$0.01} \\ 
\end{tabular} }
\end{table} 

\begin{table}[!htbp] \centering 
  \caption{Morality.and.Ethics} 
  \label{} 
\resizebox{.8\columnwidth}{!}{%

\begin{tabular}{@{\extracolsep{5pt}}lc} 
\hline \\[-1.8ex] 
 countryEU & $-$0.195$^{***}$ (0.013) \\ 
  countryGB & $-$0.495$^{***}$ (0.009) \\ 
  has\_hashtag1 & 0.169$^{***}$ (0.007) \\ 
  has\_mention1 & $-$0.231$^{***}$ (0.007) \\ 
  has\_url1 & 0.064$^{***}$ (0.009) \\ 
  is\_quote\_status1 & 0.126$^{***}$ (0.007) \\ 
  is\_reply1 & $-$0.125$^{***}$ (0.007) \\ 
  is\_verified1 & 0.164$^{***}$ (0.014) \\ 
  log\_chars & 0.874$^{***}$ (0.007) \\ 
  log\_followers & $-$0.044$^{***}$ (0.002) \\ 
  log\_following & 0.033$^{***}$ (0.002) \\ 
  log\_statuses & $-$0.006$^{***}$ (0.002) \\ 
  Constant & $-$6.794$^{***}$ (0.077) \\ 
 \hline \\[-1.8ex] 
Observations & 2,661,435 \\ 
Log Likelihood & $-$768,383.100 \\ 
Akaike Inf. Crit. & 1,536,798.000 \\ 
Bayesian Inf. Crit. & 1,537,003.000 \\ 
\hline 
\hline \\[-1.8ex] 
\textit{Note:}  & \multicolumn{1}{r}{$^{*}$p$<$0.1; $^{**}$p$<$0.05; $^{***}$p$<$0.01} \\ 
\end{tabular} }
\end{table} 

\begin{table}[!htbp] \centering 
  \caption{Policy.Prescription.and.Evaluation} 
  \label{} 
 \resizebox{.8\columnwidth}{!}{%

\begin{tabular}{@{\extracolsep{5pt}}lc} 
\hline \\[-1.8ex] 
 countryEU & $-$0.152$^{***}$ (0.008) \\ 
  countryGB & 0.356$^{***}$ (0.004) \\ 
  has\_hashtag1 & 0.220$^{***}$ (0.005) \\ 
  has\_mention1 & 0.107$^{***}$ (0.004) \\ 
  has\_url1 & 0.213$^{***}$ (0.005) \\ 
  is\_quote\_status1 & 0.059$^{***}$ (0.005) \\ 
  is\_reply1 & $-$0.083$^{***}$ (0.005) \\ 
  is\_verified1 & 0.337$^{***}$ (0.009) \\ 
  log\_chars & 0.670$^{***}$ (0.004) \\ 
  log\_followers & $-$0.046$^{***}$ (0.001) \\ 
  log\_following & 0.021$^{***}$ (0.001) \\ 
  log\_statuses & 0.042$^{***}$ (0.001) \\ 
  Constant & $-$4.886$^{***}$ (0.031) \\ 
 \hline \\[-1.8ex] 
Observations & 2,661,435 \\ 
Log Likelihood & $-$1,574,272.000 \\ 
Akaike Inf. Crit. & 3,148,576.000 \\ 
Bayesian Inf. Crit. & 3,148,780.000 \\ 
\hline 
\hline \\[-1.8ex] 
\textit{Note:}  & \multicolumn{1}{r}{$^{*}$p$<$0.1; $^{**}$p$<$0.05; $^{***}$p$<$0.01} \\ 
\end{tabular} }
\end{table} 

\begin{table}[!htbp] \centering 
  \caption{Political.Factors.and.Implications} 
  \label{} 
 \resizebox{.8\columnwidth}{!}{%

\begin{tabular}{@{\extracolsep{5pt}}lc} 
\hline \\[-1.8ex] 
 countryEU & $-$0.531$^{***}$ (0.007) \\ 
  countryGB & $-$0.326$^{***}$ (0.004) \\ 
  has\_hashtag1 & 0.192$^{***}$ (0.005) \\ 
  has\_mention1 & 0.180$^{***}$ (0.004) \\ 
  has\_url1 & 0.214$^{***}$ (0.005) \\ 
  is\_quote\_status1 & 0.156$^{***}$ (0.005) \\ 
  is\_reply1 & $-$0.301$^{***}$ (0.004) \\ 
  is\_verified1 & $-$0.060$^{***}$ (0.008) \\ 
  log\_chars & 0.984$^{***}$ (0.004) \\ 
  log\_followers & $-$0.044$^{***}$ (0.001) \\ 
  log\_following & 0.052$^{***}$ (0.001) \\ 
  log\_statuses & 0.075$^{***}$ (0.001) \\ 
  Constant & $-$6.163$^{***}$ (0.028) \\ 
 \hline \\[-1.8ex] 
Observations & 2,661,435 \\ 
Log Likelihood & $-$1,735,632.000 \\ 
Akaike Inf. Crit. & 3,471,297.000 \\ 
Bayesian Inf. Crit. & 3,471,501.000 \\ 
\hline 
\hline \\[-1.8ex] 
\textit{Note:}  & \multicolumn{1}{r}{$^{*}$p$<$0.1; $^{**}$p$<$0.05; $^{***}$p$<$0.01} \\ 
\end{tabular} }
\end{table} 

\begin{table}[!htbp] \centering 
  \caption{Public.Sentiment} 
  \label{} 
\resizebox{.8\columnwidth}{!}{%

\begin{tabular}{@{\extracolsep{5pt}}lc} 
\hline \\[-1.8ex] 
 countryEU & 0.047$^{***}$ (0.015) \\ 
  countryGB & 0.286$^{***}$ (0.008) \\ 
  has\_hashtag1 & 0.278$^{***}$ (0.009) \\ 
  has\_mention1 & $-$0.017$^{*}$ (0.009) \\ 
  has\_url1 & 0.462$^{***}$ (0.013) \\ 
  is\_quote\_status1 & $-$0.135$^{***}$ (0.010) \\ 
  is\_reply1 & $-$0.279$^{***}$ (0.009) \\ 
  is\_verified1 & 0.049$^{***}$ (0.017) \\ 
  log\_chars & 0.538$^{***}$ (0.008) \\ 
  log\_followers & 0.011$^{***}$ (0.003) \\ 
  log\_following & 0.009$^{***}$ (0.003) \\ 
  log\_statuses & 0.0004 (0.002) \\ 
  Constant & $-$6.210$^{***}$ (0.085) \\ 
 \hline \\[-1.8ex] 
Observations & 2,661,435 \\ 
Log Likelihood & $-$523,820.000 \\ 
Akaike Inf. Crit. & 1,047,672.000 \\ 
Bayesian Inf. Crit. & 1,047,877.000 \\ 
\hline 
\hline \\[-1.8ex] 
\textit{Note:}  & \multicolumn{1}{r}{$^{*}$p$<$0.1; $^{**}$p$<$0.05; $^{***}$p$<$0.01} \\ 
\end{tabular} }
\end{table} 

\begin{table}[!htbp] \centering 
  \caption{Quality.of.Life} 
  \label{} 
 \resizebox{.8\columnwidth}{!}{%

\begin{tabular}{@{\extracolsep{5pt}}lc} 
\hline \\[-1.8ex] 
 countryEU & 0.059$^{***}$ (0.016) \\ 
  countryGB & 0.168$^{***}$ (0.009) \\ 
  has\_hashtag1 & $-$0.178$^{***}$ (0.011) \\ 
  has\_mention1 & $-$0.565$^{***}$ (0.010) \\ 
  has\_url1 & 0.139$^{***}$ (0.014) \\ 
  is\_quote\_status1 & 0.131$^{***}$ (0.010) \\ 
  is\_reply1 & 0.349$^{***}$ (0.010) \\ 
  is\_verified1 & 0.123$^{***}$ (0.018) \\ 
  log\_chars & 1.050$^{***}$ (0.009) \\ 
  log\_followers & 0.033$^{***}$ (0.003) \\ 
  log\_following & $-$0.031$^{***}$ (0.003) \\ 
  log\_statuses & $-$0.063$^{***}$ (0.002) \\ 
  Constant & $-$7.958$^{***}$ (0.058) \\ 
 \hline \\[-1.8ex] 
Observations & 2,661,435 \\ 
Log Likelihood & $-$491,143.800 \\ 
Akaike Inf. Crit. & 982,319.500 \\ 
Bayesian Inf. Crit. & 982,524.300 \\ 
\hline 
\hline \\[-1.8ex] 
\textit{Note:}  & \multicolumn{1}{r}{$^{*}$p$<$0.1; $^{**}$p$<$0.05; $^{***}$p$<$0.01} \\ 
\end{tabular} }
\end{table} 

\begin{table}[!htbp] \centering 
  \caption{Security.and.Defense} 
  \label{} 
 \resizebox{.8\columnwidth}{!}{%

\begin{tabular}{@{\extracolsep{5pt}}lc} 

\hline \\[-1.8ex] 
 countryEU & $-$0.260$^{***}$ (0.009) \\ 
  countryGB & $-$0.469$^{***}$ (0.006) \\ 
  has\_hashtag1 & 0.203$^{***}$ (0.005) \\ 
  has\_mention1 & 0.012$^{**}$ (0.005) \\ 
  has\_url1 & 0.260$^{***}$ (0.006) \\ 
  is\_quote\_status1 & 0.006 (0.005) \\ 
  is\_reply1 & $-$0.104$^{***}$ (0.005) \\ 
  is\_verified1 & $-$0.147$^{***}$ (0.010) \\ 
  log\_chars & 0.602$^{***}$ (0.005) \\ 
  log\_followers & $-$0.016$^{***}$ (0.002) \\ 
  log\_following & 0.014$^{***}$ (0.002) \\ 
  log\_statuses & 0.026$^{***}$ (0.001) \\ 
  Constant & $-$4.981$^{***}$ (0.071) \\ 
 \hline \\[-1.8ex] 
Observations & 2,661,435 \\ 
Log Likelihood & $-$1,332,280.000 \\ 
Akaike Inf. Crit. & 2,664,591.000 \\ 
Bayesian Inf. Crit. & 2,664,796.000 \\ 
\hline 
\hline \\[-1.8ex] 
\textit{Note:}  & \multicolumn{1}{r}{$^{*}$p$<$0.1; $^{**}$p$<$0.05; $^{***}$p$<$0.01} \\ 
\end{tabular} }
\end{table}

\begin{table}[!htbp] \centering 
  \caption{Hero..Cultural.Diversity} 
  \label{} 
\resizebox{.8\columnwidth}{!}{%

\begin{tabular}{@{\extracolsep{5pt}}lc} 

\hline \\[-1.8ex] 
 countryEU & 0.041 (0.031) \\ 
  countryGB & 0.051$^{***}$ (0.018) \\ 
  has\_hashtag1 & 0.114$^{***}$ (0.020) \\ 
  has\_mention1 & $-$0.208$^{***}$ (0.020) \\ 
  has\_url1 & $-$0.480$^{***}$ (0.021) \\ 
  is\_quote\_status1 & 0.270$^{***}$ (0.020) \\ 
  is\_reply1 & 0.200$^{***}$ (0.020) \\ 
  is\_verified1 & 0.405$^{***}$ (0.035) \\ 
  log\_chars & 0.449$^{***}$ (0.017) \\ 
  log\_followers & $-$0.003 (0.006) \\ 
  log\_following & 0.004 (0.006) \\ 
  log\_statuses & $-$0.095$^{***}$ (0.004) \\ 
  Constant & $-$5.560$^{***}$ (0.104) \\ 
 \hline \\[-1.8ex] 
Observations & 2,661,435 \\ 
Log Likelihood & $-$164,157.100 \\ 
Akaike Inf. Crit. & 328,346.100 \\ 
Bayesian Inf. Crit. & 328,550.800 \\ 
\hline 
\hline \\[-1.8ex] 
\textit{Note:}  & \multicolumn{1}{r}{$^{*}$p$<$0.1; $^{**}$p$<$0.05; $^{***}$p$<$0.01} \\ 
\end{tabular} }
\end{table} 

\begin{table}[!htbp] \centering 
  \caption{Hero..Integration} 
  \label{} 
\resizebox{.8\columnwidth}{!}{%

\begin{tabular}{@{\extracolsep{5pt}}lc} 

\hline \\[-1.8ex] 
 countryEU & $-$0.096$^{***}$ (0.023) \\ 
  countryGB & $-$0.073$^{***}$ (0.013) \\ 
  has\_hashtag1 & $-$0.103$^{***}$ (0.015) \\ 
  has\_mention1 & $-$0.399$^{***}$ (0.014) \\ 
  has\_url1 & $-$0.169$^{***}$ (0.016) \\ 
  is\_quote\_status1 & 0.102$^{***}$ (0.014) \\ 
  is\_reply1 & 0.199$^{***}$ (0.014) \\ 
  is\_verified1 & 0.404$^{***}$ (0.023) \\ 
  log\_chars & 0.649$^{***}$ (0.012) \\ 
  log\_followers & 0.031$^{***}$ (0.004) \\ 
  log\_following & $-$0.033$^{***}$ (0.004) \\ 
  log\_statuses & $-$0.088$^{***}$ (0.003) \\ 
  Constant & $-$6.012$^{***}$ (0.075) \\ 
 \hline \\[-1.8ex] 
Observations & 2,661,435 \\ 
Log Likelihood & $-$287,803.800 \\ 
Akaike Inf. Crit. & 575,639.600 \\ 
Bayesian Inf. Crit. & 575,844.400 \\ 
\hline 
\hline \\[-1.8ex] 
\textit{Note:}  & \multicolumn{1}{r}{$^{*}$p$<$0.1; $^{**}$p$<$0.05; $^{***}$p$<$0.01} \\ 
\end{tabular} }
\end{table} 

\begin{table}[!htbp] \centering 
  \caption{Hero..Worker} 
  \label{} 

\resizebox{.8\columnwidth}{!}{%

\begin{tabular}{@{\extracolsep{5pt}}lc} 
\hline \\[-1.8ex] 
 countryEU & 0.287$^{***}$ (0.027) \\ 
  countryGB & 1.287$^{***}$ (0.011) \\ 
  has\_hashtag1 & $-$0.134$^{***}$ (0.020) \\ 
  has\_mention1 & $-$0.123$^{***}$ (0.018) \\ 
  has\_url1 & 0.119$^{***}$ (0.020) \\ 
  is\_quote\_status1 & 0.111$^{***}$ (0.019) \\ 
  is\_reply1 & 0.501$^{***}$ (0.018) \\ 
  is\_verified1 & 0.342$^{***}$ (0.033) \\ 
  log\_chars & 0.462$^{***}$ (0.015) \\ 
  log\_followers & $-$0.049$^{***}$ (0.005) \\ 
  log\_following & 0.009$^{*}$ (0.005) \\ 
  log\_statuses & $-$0.016$^{***}$ (0.004) \\ 
  Constant & $-$6.720$^{***}$ (0.099) \\ 
 \hline \\[-1.8ex] 
Observations & 2,661,435 \\ 
Log Likelihood & $-$210,181.700 \\ 
Akaike Inf. Crit. & 420,395.400 \\ 
Bayesian Inf. Crit. & 420,600.100 \\ 
\hline 
\hline \\[-1.8ex] 
\textit{Note:}  & \multicolumn{1}{r}{$^{*}$p$<$0.1; $^{**}$p$<$0.05; $^{***}$p$<$0.01} \\ 
\end{tabular} }
\end{table} 

\begin{table}[!htbp] \centering 
  \caption{Threat..Fiscal} 
  \label{} 
 \resizebox{.8\columnwidth}{!}{%

\begin{tabular}{@{\extracolsep{5pt}}lc} 

\hline \\[-1.8ex] 
 countryEU & $-$0.633$^{***}$ (0.017) \\ 
  countryGB & $-$0.435$^{***}$ (0.009) \\ 
  has\_hashtag1 & $-$0.244$^{***}$ (0.010) \\ 
  has\_mention1 & 0.227$^{***}$ (0.008) \\ 
  has\_url1 & $-$0.238$^{***}$ (0.009) \\ 
  is\_quote\_status1 & 0.716$^{***}$ (0.008) \\ 
  is\_reply1 & 0.557$^{***}$ (0.009) \\ 
  is\_verified1 & $-$1.728$^{***}$ (0.034) \\ 
  log\_chars & 0.480$^{***}$ (0.007) \\ 
  log\_followers & 0.019$^{***}$ (0.002) \\ 
  log\_following & 0.023$^{***}$ (0.003) \\ 
  log\_statuses & $-$0.035$^{***}$ (0.002) \\ 
  Constant & $-$5.272$^{***}$ (0.085) \\ 
 \hline \\[-1.8ex] 
Observations & 2,661,435 \\ 
Log Likelihood & $-$709,024.600 \\ 
Akaike Inf. Crit. & 1,418,081.000 \\ 
Bayesian Inf. Crit. & 1,418,286.000 \\ 
\hline 
\hline \\[-1.8ex] 
\textit{Note:}  & \multicolumn{1}{r}{$^{*}$p$<$0.1; $^{**}$p$<$0.05; $^{***}$p$<$0.01} \\ 
\end{tabular} }
\end{table} 

\begin{table}[!htbp] \centering 
  \caption{Threat..Jobs} 
  \label{} 
 \resizebox{.8\columnwidth}{!}{%

\begin{tabular}{@{\extracolsep{5pt}}lc} 

\hline \\[-1.8ex] 
 countryEU & $-$0.339$^{***}$ (0.043) \\ 
  countryGB & 0.438$^{***}$ (0.017) \\ 
  has\_hashtag1 & $-$0.233$^{***}$ (0.028) \\ 
  has\_mention1 & $-$0.158$^{***}$ (0.024) \\ 
  has\_url1 & $-$0.220$^{***}$ (0.025) \\ 
  is\_quote\_status1 & 0.488$^{***}$ (0.024) \\ 
  is\_reply1 & 0.710$^{***}$ (0.024) \\ 
  is\_verified1 & $-$0.702$^{***}$ (0.071) \\ 
  log\_chars & 0.560$^{***}$ (0.019) \\ 
  log\_followers & $-$0.090$^{***}$ (0.007) \\ 
  log\_following & $-$0.013$^{**}$ (0.007) \\ 
  log\_statuses & 0.046$^{***}$ (0.005) \\ 
  Constant & $-$7.588$^{***}$ (0.101) \\ 
 \hline \\[-1.8ex] 
Observations & 2,661,435 \\ 
Log Likelihood & $-$143,493.400 \\ 
Akaike Inf. Crit. & 287,018.800 \\ 
Bayesian Inf. Crit. & 287,223.500 \\ 
\hline 
\hline \\[-1.8ex] 
\textit{Note:}  & \multicolumn{1}{r}{$^{*}$p$<$0.1; $^{**}$p$<$0.05; $^{***}$p$<$0.01} \\ 
\end{tabular} }
\end{table} 

\begin{table}[!htbp] \centering 
  \caption{Threat..National.Cohesion} 
  \label{} 
\resizebox{.8\columnwidth}{!}{%

\begin{tabular}{@{\extracolsep{5pt}}lc} 

\hline \\[-1.8ex] 
 countryEU & 0.279$^{***}$ (0.012) \\ 
  countryGB & 0.083$^{***}$ (0.008) \\ 
  has\_hashtag1 & 0.117$^{***}$ (0.009) \\ 
  has\_mention1 & 0.067$^{***}$ (0.009) \\ 
  has\_url1 & $-$0.372$^{***}$ (0.009) \\ 
  is\_quote\_status1 & 0.863$^{***}$ (0.008) \\ 
  is\_reply1 & 0.631$^{***}$ (0.009) \\ 
  is\_verified1 & $-$1.817$^{***}$ (0.035) \\ 
  log\_chars & 0.643$^{***}$ (0.007) \\ 
  log\_followers & 0.040$^{***}$ (0.003) \\ 
  log\_following & $-$0.003 (0.003) \\ 
  log\_statuses & $-$0.028$^{***}$ (0.002) \\ 
  Constant & $-$6.122$^{***}$ (0.061) \\ 
 \hline \\[-1.8ex] 
Observations & 2,661,435 \\ 
Log Likelihood & $-$659,886.300 \\ 
Akaike Inf. Crit. & 1,319,805.000 \\ 
Bayesian Inf. Crit. & 1,320,009.000 \\ 
\hline 
\hline \\[-1.8ex] 
\textit{Note:}  & \multicolumn{1}{r}{$^{*}$p$<$0.1; $^{**}$p$<$0.05; $^{***}$p$<$0.01} \\ 
\end{tabular} }
\end{table} 

\begin{table}[!htbp] \centering 
  \caption{Threat..Public.Order} 
  \label{} 
\resizebox{.8\columnwidth}{!}{%

\begin{tabular}{@{\extracolsep{5pt}}lc} 

\hline \\[-1.8ex] 
 countryEU & $-$0.675$^{***}$ (0.009) \\ 
  countryGB & $-$1.381$^{***}$ (0.007) \\ 
  has\_hashtag1 & 0.053$^{***}$ (0.005) \\ 
  has\_mention1 & 0.216$^{***}$ (0.005) \\ 
  has\_url1 & 0.029$^{***}$ (0.005) \\ 
  is\_quote\_status1 & 0.448$^{***}$ (0.005) \\ 
  is\_reply1 & 0.244$^{***}$ (0.005) \\ 
  is\_verified1 & $-$1.226$^{***}$ (0.013) \\ 
  log\_chars & 0.274$^{***}$ (0.004) \\ 
  log\_followers & 0.006$^{***}$ (0.001) \\ 
  log\_following & 0.045$^{***}$ (0.002) \\ 
  log\_statuses & $-$0.012$^{***}$ (0.001) \\ 
  Constant & $-$2.966$^{***}$ (0.056) \\ 
 \hline \\[-1.8ex] 
Observations & 2,661,435 \\ 
Log Likelihood & $-$1,469,478.000 \\ 
Akaike Inf. Crit. & 2,938,987.000 \\ 
Bayesian Inf. Crit. & 2,939,192.000 \\ 
\hline 
\hline \\[-1.8ex] 
\textit{Note:}  & \multicolumn{1}{r}{$^{*}$p$<$0.1; $^{**}$p$<$0.05; $^{***}$p$<$0.01} \\ 
\end{tabular} }
\end{table}

\begin{table}[!htbp] \centering 
  \caption{Victim..Discrimination} 
  \label{} 
 \resizebox{.8\columnwidth}{!}{%

\begin{tabular}{@{\extracolsep{5pt}}lc} 
\hline \\[-1.8ex] 
 countryEU & 0.046$^{***}$ (0.012) \\ 
  countryGB & 0.453$^{***}$ (0.006) \\ 
  has\_hashtag1 & $-$0.165$^{***}$ (0.009) \\ 
  has\_mention1 & $-$0.162$^{***}$ (0.007) \\ 
  has\_url1 & $-$0.103$^{***}$ (0.009) \\ 
  is\_quote\_status1 & 0.201$^{***}$ (0.008) \\ 
  is\_reply1 & 0.173$^{***}$ (0.008) \\ 
  is\_verified1 & 0.473$^{***}$ (0.015) \\ 
  log\_chars & 1.022$^{***}$ (0.007) \\ 
  log\_followers & $-$0.151$^{***}$ (0.002) \\ 
  log\_following & 0.074$^{***}$ (0.002) \\ 
  log\_statuses & 0.101$^{***}$ (0.002) \\ 
  Constant & $-$8.099$^{***}$ (0.058) \\ 
 \hline \\[-1.8ex] 
Observations & 2,661,435 \\ 
Log Likelihood & $-$770,001.000 \\ 
Akaike Inf. Crit. & 1,540,034.000 \\ 
Bayesian Inf. Crit. & 1,540,239.000 \\ 
\hline 
\hline \\[-1.8ex] 
\textit{Note:}  & \multicolumn{1}{r}{$^{*}$p$<$0.1; $^{**}$p$<$0.05; $^{***}$p$<$0.01} \\ 
\end{tabular} }
\end{table} 

\begin{table}[!htbp] \centering 
  \caption{Victim..Global.Economy} 
  \label{} 
\resizebox{.8\columnwidth}{!}{%

\begin{tabular}{@{\extracolsep{5pt}}lc} 
\hline \\[-1.8ex] 
 countryEU & 0.914$^{***}$ (0.033) \\ 
  countryGB & 1.088$^{***}$ (0.019) \\ 
  has\_hashtag1 & $-$0.344$^{***}$ (0.036) \\ 
  has\_mention1 & 0.018 (0.029) \\ 
  has\_url1 & 0.015 (0.033) \\ 
  is\_quote\_status1 & 0.213$^{***}$ (0.031) \\ 
  is\_reply1 & 0.387$^{***}$ (0.029) \\ 
  is\_verified1 & 0.272$^{***}$ (0.057) \\ 
  log\_chars & 0.603$^{***}$ (0.024) \\ 
  log\_followers & $-$0.059$^{***}$ (0.008) \\ 
  log\_following & $-$0.004 (0.008) \\ 
  log\_statuses & $-$0.015$^{**}$ (0.006) \\ 
  Constant & $-$8.433$^{***}$ (0.130) \\ 
 \hline \\[-1.8ex] 
Observations & 2,661,435 \\ 
Log Likelihood & $-$94,328.800 \\ 
Akaike Inf. Crit. & 188,689.600 \\ 
Bayesian Inf. Crit. & 188,894.300 \\ 
\hline 
\hline \\[-1.8ex] 
\textit{Note:}  & \multicolumn{1}{r}{$^{*}$p$<$0.1; $^{**}$p$<$0.05; $^{***}$p$<$0.01} \\ 
\end{tabular} }
\end{table} 

\begin{table}[!htbp] \centering 
  \caption{Victim..Humanitarian} 
  \label{} 
 \resizebox{.8\columnwidth}{!}{%

\begin{tabular}{@{\extracolsep{5pt}}lc} 
\hline \\[-1.8ex] 
 countryEU & $-$0.059$^{***}$ (0.013) \\ 
  countryGB & $-$0.453$^{***}$ (0.009) \\ 
  has\_hashtag1 & 0.198$^{***}$ (0.007) \\ 
  has\_mention1 & $-$0.339$^{***}$ (0.007) \\ 
  has\_url1 & 0.534$^{***}$ (0.010) \\ 
  is\_quote\_status1 & $-$0.306$^{***}$ (0.008) \\ 
  is\_reply1 & $-$0.570$^{***}$ (0.008) \\ 
  is\_verified1 & 0.553$^{***}$ (0.013) \\ 
  log\_chars & 0.473$^{***}$ (0.007) \\ 
  log\_followers & $-$0.048$^{***}$ (0.002) \\ 
  log\_following & 0.033$^{***}$ (0.002) \\ 
  log\_statuses & 0.045$^{***}$ (0.002) \\ 
  Constant & $-$5.456$^{***}$ (0.098) \\ 
 \hline \\[-1.8ex] 
Observations & 2,661,435 \\ 
Log Likelihood & $-$679,409.600 \\ 
Akaike Inf. Crit. & 1,358,851.000 \\ 
Bayesian Inf. Crit. & 1,359,056.000 \\ 
\hline 
\hline \\[-1.8ex] 
\textit{Note:}  & \multicolumn{1}{r}{$^{*}$p$<$0.1; $^{**}$p$<$0.05; $^{***}$p$<$0.01} \\ 
\end{tabular} }
\end{table} 

\begin{table}[!htbp] \centering 
  \caption{Victim..War} 
  \label{}
 \resizebox{.8\columnwidth}{!}{%

\begin{tabular}{@{\extracolsep{5pt}}lc} 
\hline \\[-1.8ex] 
 countryEU & 0.631$^{***}$ (0.046) \\ 
  countryGB & 0.228$^{***}$ (0.033) \\ 
  has\_hashtag1 & $-$0.135$^{***}$ (0.042) \\ 
  has\_mention1 & $-$0.596$^{***}$ (0.036) \\ 
  has\_url1 & 0.529$^{***}$ (0.058) \\ 
  is\_quote\_status1 & $-$0.144$^{***}$ (0.039) \\ 
  is\_reply1 & 0.274$^{***}$ (0.038) \\ 
  is\_verified1 & 0.612$^{***}$ (0.060) \\ 
  log\_chars & 1.017$^{***}$ (0.033) \\ 
  log\_followers & $-$0.062$^{***}$ (0.010) \\ 
  log\_following & 0.015 (0.011) \\ 
  log\_statuses & 0.005 (0.008) \\ 
  Constant & $-$11.326$^{***}$ (0.175) \\ 
 \hline \\[-1.8ex] 
Observations & 2,661,435 \\ 
Log Likelihood & $-$56,045.610 \\ 
Akaike Inf. Crit. & 112,123.200 \\ 
Bayesian Inf. Crit. & 112,327.900 \\ 
\hline 
\hline \\[-1.8ex] 
\textit{Note:}  & \multicolumn{1}{r}{$^{*}$p$<$0.1; $^{**}$p$<$0.05; $^{***}$p$<$0.01} \\ 
\end{tabular} }
\end{table} 

\begin{table}[!htbp] \centering 
  \caption{Episodic} 
  \label{} 
\resizebox{.8\columnwidth}{!}{%

\begin{tabular}{@{\extracolsep{5pt}}lc} 
\hline \\[-1.8ex] 
 countryEU & $-$0.147$^{***}$ (0.007) \\ 
  countryGB & $-$0.204$^{***}$ (0.004) \\ 
  has\_hashtag1 & 0.017$^{***}$ (0.005) \\ 
  has\_mention1 & 0.174$^{***}$ (0.005) \\ 
  has\_url1 & 0.451$^{***}$ (0.005) \\ 
  is\_quote\_status1 & $-$0.866$^{***}$ (0.005) \\ 
  is\_reply1 & $-$1.078$^{***}$ (0.005) \\ 
  is\_verified1 & 0.578$^{***}$ (0.011) \\ 
  log\_chars & $-$0.288$^{***}$ (0.004) \\ 
  log\_followers & $-$0.006$^{***}$ (0.001) \\ 
  log\_following & 0.024$^{***}$ (0.001) \\ 
  log\_statuses & 0.043$^{***}$ (0.001) \\ 
  Constant & 1.857$^{***}$ (0.032) \\ 
 \hline \\[-1.8ex] 
Observations & 2,661,435 \\ 
Log Likelihood & $-$1,633,108.000 \\ 
Akaike Inf. Crit. & 3,266,249.000 \\ 
Bayesian Inf. Crit. & 3,266,453.000 \\ 
\hline 
\hline \\[-1.8ex] 
\textit{Note:}  & \multicolumn{1}{r}{$^{*}$p$<$0.1; $^{**}$p$<$0.05; $^{***}$p$<$0.01} \\ 
\end{tabular} }
\end{table} 

\begin{table}[!htbp] \centering 
  \caption{Thematic} 
  \label{tab:reg_country_end} 

 \resizebox{.8\columnwidth}{!}{%

\begin{tabular}{@{\extracolsep{5pt}}lc} 
\hline \\[-1.8ex] 
 countryEU & $-$0.249$^{***}$ (0.008) \\ 
  countryGB & $-$0.012$^{***}$ (0.005) \\ 
  has\_hashtag1 & $-$0.273$^{***}$ (0.005) \\ 
  has\_mention1 & $-$0.062$^{***}$ (0.004) \\ 
  has\_url1 & $-$0.471$^{***}$ (0.005) \\ 
  is\_quote\_status1 & 1.094$^{***}$ (0.005) \\ 
  is\_reply1 & 1.061$^{***}$ (0.005) \\ 
  is\_verified1 & $-$0.408$^{***}$ (0.009) \\ 
  log\_chars & 1.845$^{***}$ (0.004) \\ 
  log\_followers & $-$0.047$^{***}$ (0.001) \\ 
  log\_following & 0.022$^{***}$ (0.001) \\ 
  log\_statuses & $-$0.018$^{***}$ (0.001) \\ 
  Constant & $-$9.106$^{***}$ (0.036) \\ 
 \hline \\[-1.8ex] 
Observations & 2,661,435 \\ 
Log Likelihood & $-$1,489,756.000 \\ 
Akaike Inf. Crit. & 2,979,544.000 \\ 
Bayesian Inf. Crit. & 2,979,749.000 \\ 
\hline 
\hline \\[-1.8ex] 
\textit{Note:}  & \multicolumn{1}{r}{$^{*}$p$<$0.1; $^{**}$p$<$0.05; $^{***}$p$<$0.01} \\ 
\end{tabular} }
\end{table} 

\section{Frame-building (ideology) regressions}
\label{app:ideology-regressions}

Tables \ref{tab:reg_ideology_start}-\ref{tab:reg_ideology_end} show independent variable coefficients for logit regressions predicting frames from ideology.
\vspace{-.5em}

\begin{table}[!htbp] \centering 
  \caption{Capacity.and.Resources} 
  \vspace{-1em}

    \label{tab:reg_ideology_start} 
\resizebox{.8\columnwidth}{!}{%

\begin{tabular}{@{\extracolsep{5pt}}lc} 
\hline \\[-1.8ex] 
 has\_hashtag1 & $-$0.268$^{***}$ (0.014) \\ 
  has\_mention1 & $-$0.056$^{***}$ (0.011) \\ 
  has\_url1 & $-$0.049$^{***}$ (0.013) \\ 
  is\_quote\_status1 & 0.248$^{***}$ (0.012) \\ 
  is\_reply1 & 0.329$^{***}$ (0.011) \\ 
  is\_verified1 & $-$0.084$^{***}$ (0.025) \\ 
  log\_chars & 0.767$^{***}$ (0.010) \\ 
  log\_followers & $-$0.013$^{***}$ (0.004) \\ 
  log\_following & $-$0.019$^{***}$ (0.004) \\ 
  log\_statuses & $-$0.023$^{***}$ (0.003) \\ 
  ideology & 0.178$^{***}$ (0.002) \\ 
  Constant & $-$6.763$^{***}$ (0.101) \\ 
 \hline \\[-1.8ex] 
Observations & 1,680,164 \\ 
Log Likelihood & $-$373,804.500 \\ 
Akaike Inf. Crit. & 747,638.900 \\ 
Bayesian Inf. Crit. & 747,823.900 \\ 
\hline 
\hline \\[-1.8ex] 
\textit{Note:}  & \multicolumn{1}{r}{$^{*}$p$<$0.1; $^{**}$p$<$0.05; $^{***}$p$<$0.01} \\ 
\end{tabular} }
\end{table} 

\vspace{-1em}

\begin{table}[!htbp] \centering 
  \caption{Crime.and.Punishment}
  \vspace{-1em}

  \label{} 

\resizebox{.8\columnwidth}{!}{%

\begin{tabular}{@{\extracolsep{5pt}}lc} 

\hline \\[-1.8ex] 
 has\_hashtag1 & $-$0.074$^{***}$ (0.006) \\ 
  has\_mention1 & $-$0.145$^{***}$ (0.005) \\ 
  has\_url1 & 0.152$^{***}$ (0.006) \\ 
  is\_quote\_status1 & 0.009 (0.006) \\ 
  is\_reply1 & 0.096$^{***}$ (0.006) \\ 
  is\_verified1 & $-$0.213$^{***}$ (0.011) \\ 
  log\_chars & 0.302$^{***}$ (0.005) \\ 
  log\_followers & $-$0.010$^{***}$ (0.002) \\ 
  log\_following & 0.010$^{***}$ (0.002) \\ 
  log\_statuses & 0.021$^{***}$ (0.001) \\ 
  ideology & 0.174$^{***}$ (0.001) \\ 
  Constant & $-$2.652$^{***}$ (0.042) \\ 
 \hline \\[-1.8ex] 
Observations & 1,680,164 \\ 
Log Likelihood & $-$1,052,324.000 \\ 
Akaike Inf. Crit. & 2,104,679.000 \\ 
Bayesian Inf. Crit. & 2,104,864.000 \\ 
\hline 
\hline \\[-1.8ex] 
\textit{Note:}  & \multicolumn{1}{r}{$^{*}$p$<$0.1; $^{**}$p$<$0.05; $^{***}$p$<$0.01} \\ 
\end{tabular} }
\end{table} 

\vspace{-1em}

\begin{table}[!htbp] \centering 
  \caption{Cultural.Identity} 
  \vspace{-1em}

  \label{} 
\resizebox{.8\columnwidth}{!}{%

\begin{tabular}{@{\extracolsep{5pt}}lc} 
\hline \\[-1.8ex] 
 has\_hashtag1 & $-$0.104$^{***}$ (0.008) \\ 
  has\_mention1 & $-$0.336$^{***}$ (0.008) \\ 
  has\_url1 & $-$0.311$^{***}$ (0.009) \\ 
  is\_quote\_status1 & 0.022$^{***}$ (0.008) \\ 
  is\_reply1 & 0.253$^{***}$ (0.008) \\ 
  is\_verified1 & $-$0.043$^{***}$ (0.014) \\ 
  log\_chars & 0.725$^{***}$ (0.007) \\ 
  log\_followers & 0.025$^{***}$ (0.002) \\ 
  log\_following & $-$0.005$^{**}$ (0.003) \\ 
  log\_statuses & $-$0.021$^{***}$ (0.002) \\ 
  ideology & $-$0.097$^{***}$ (0.001) \\ 
  Constant & $-$4.993$^{***}$ (0.051) \\ 
 \hline \\[-1.8ex] 
Observations & 1,680,164 \\ 
Log Likelihood & $-$654,480.300 \\ 
Akaike Inf. Crit. & 1,308,991.000 \\ 
Bayesian Inf. Crit. & 1,309,176.000 \\ 
\hline 
\hline \\[-1.8ex] 
\textit{Note:}  & \multicolumn{1}{r}{$^{*}$p$<$0.1; $^{**}$p$<$0.05; $^{***}$p$<$0.01} \\ 
\end{tabular} }
\end{table} 

\begin{table}[!htbp] \centering 
  \caption{Economic} 
  \label{} 
\resizebox{.8\columnwidth}{!}{%

\begin{tabular}{@{\extracolsep{5pt}}lc} 
\hline \\[-1.8ex] 
 has\_hashtag1 & $-$0.140$^{***}$ (0.007) \\ 
  has\_mention1 & $-$0.003 (0.006) \\ 
  has\_url1 & $-$0.067$^{***}$ (0.008) \\ 
  is\_quote\_status1 & 0.181$^{***}$ (0.006) \\ 
  is\_reply1 & 0.138$^{***}$ (0.006) \\ 
  is\_verified1 & $-$0.043$^{***}$ (0.013) \\ 
  log\_chars & 0.973$^{***}$ (0.006) \\ 
  log\_followers & $-$0.069$^{***}$ (0.002) \\ 
  log\_following & 0.025$^{***}$ (0.002) \\ 
  log\_statuses & 0.041$^{***}$ (0.002) \\ 
  ideology & 0.048$^{***}$ (0.001) \\ 
  Constant & $-$6.375$^{***}$ (0.066) \\ 
 \hline \\[-1.8ex] 
Observations & 1,680,164 \\ 
Log Likelihood & $-$882,709.100 \\ 
Akaike Inf. Crit. & 1,765,448.000 \\ 
Bayesian Inf. Crit. & 1,765,633.000 \\ 
\hline 
\hline \\[-1.8ex] 
\textit{Note:}  & \multicolumn{1}{r}{$^{*}$p$<$0.1; $^{**}$p$<$0.05; $^{***}$p$<$0.01} \\ 
\end{tabular} }
\end{table}

\begin{table}[!htbp] \centering 
  \caption{External.Regulation.and.Reputation} 
  \label{}
\resizebox{.8\columnwidth}{!}{%

\begin{tabular}{@{\extracolsep{5pt}}lc} 
\hline \\[-1.8ex] 
 has\_hashtag1 & 0.097$^{***}$ (0.014) \\ 
  has\_mention1 & $-$0.082$^{***}$ (0.013) \\ 
  has\_url1 & 0.634$^{***}$ (0.020) \\ 
  is\_quote\_status1 & $-$0.100$^{***}$ (0.014) \\ 
  is\_reply1 & $-$0.082$^{***}$ (0.013) \\ 
  is\_verified1 & $-$0.068$^{**}$ (0.027) \\ 
  log\_chars & 0.586$^{***}$ (0.012) \\ 
  log\_followers & 0.004 (0.004) \\ 
  log\_following & $-$0.031$^{***}$ (0.005) \\ 
  log\_statuses & 0.035$^{***}$ (0.003) \\ 
  ideology & 0.067$^{***}$ (0.003) \\ 
  Constant & $-$7.099$^{***}$ (0.092) \\ 
 \hline \\[-1.8ex] 
Observations & 1,680,164 \\ 
Log Likelihood & $-$254,086.900 \\ 
Akaike Inf. Crit. & 508,203.700 \\ 
Bayesian Inf. Crit. & 508,388.700 \\ 
\hline 
\hline \\[-1.8ex] 
\textit{Note:}  & \multicolumn{1}{r}{$^{*}$p$<$0.1; $^{**}$p$<$0.05; $^{***}$p$<$0.01} \\ 
\end{tabular} }
\end{table} 

\begin{table}[!htbp] \centering 
  \caption{Fairness.and.Equality} 
  \label{} 
\resizebox{.8\columnwidth}{!}{%

\begin{tabular}{@{\extracolsep{5pt}}lc} 
\hline \\[-1.8ex] 
 has\_hashtag1 & 0.065$^{***}$ (0.009) \\ 
  has\_mention1 & $-$0.126$^{***}$ (0.008) \\ 
  has\_url1 & $-$0.195$^{***}$ (0.010) \\ 
  is\_quote\_status1 & 0.564$^{***}$ (0.008) \\ 
  is\_reply1 & 0.562$^{***}$ (0.008) \\ 
  is\_verified1 & $-$0.215$^{***}$ (0.016) \\ 
  log\_chars & 0.909$^{***}$ (0.007) \\ 
  log\_followers & 0.008$^{***}$ (0.003) \\ 
  log\_following & 0.001 (0.003) \\ 
  log\_statuses & $-$0.021$^{***}$ (0.002) \\ 
  ideology & $-$0.169$^{***}$ (0.001) \\ 
  Constant & $-$6.508$^{***}$ (0.050) \\ 
 \hline \\[-1.8ex] 
Observations & 1,680,164 \\ 
Log Likelihood & $-$615,207.800 \\ 
Akaike Inf. Crit. & 1,230,446.000 \\ 
Bayesian Inf. Crit. & 1,230,631.000 \\ 
\hline 
\hline \\[-1.8ex] 
\textit{Note:}  & \multicolumn{1}{r}{$^{*}$p$<$0.1; $^{**}$p$<$0.05; $^{***}$p$<$0.01} \\ 
\end{tabular} }
\end{table} 

\begin{table}[!htbp] \centering 
  \caption{Health.and.Safety} 
  \label{} 
\resizebox{.8\columnwidth}{!}{%

\begin{tabular}{@{\extracolsep{5pt}}lc} 
\hline \\[-1.8ex] 
 has\_hashtag1 & $-$0.056$^{***}$ (0.007) \\ 
  has\_mention1 & $-$0.174$^{***}$ (0.006) \\ 
  has\_url1 & 0.500$^{***}$ (0.008) \\ 
  is\_quote\_status1 & $-$0.180$^{***}$ (0.007) \\ 
  is\_reply1 & $-$0.118$^{***}$ (0.006) \\ 
  is\_verified1 & $-$0.123$^{***}$ (0.013) \\ 
  log\_chars & 0.429$^{***}$ (0.006) \\ 
  log\_followers & $-$0.013$^{***}$ (0.002) \\ 
  log\_following & 0.025$^{***}$ (0.002) \\ 
  log\_statuses & 0.022$^{***}$ (0.002) \\ 
  ideology & 0.027$^{***}$ (0.001) \\ 
  Constant & $-$4.194$^{***}$ (0.086) \\ 
 \hline \\[-1.8ex] 
Observations & 1,680,164 \\ 
Log Likelihood & $-$814,400.700 \\ 
Akaike Inf. Crit. & 1,628,831.000 \\ 
Bayesian Inf. Crit. & 1,629,016.000 \\ 
\hline 
\hline \\[-1.8ex] 
\textit{Note:}  & \multicolumn{1}{r}{$^{*}$p$<$0.1; $^{**}$p$<$0.05; $^{***}$p$<$0.01} \\ 
\end{tabular} }
\end{table}

\begin{table}[!htbp] \centering 
  \caption{Legality..Constitutionality..Jurisdiction} 
  \label{} 
\resizebox{.8\columnwidth}{!}{%

\begin{tabular}{@{\extracolsep{5pt}}lc} 
\hline \\[-1.8ex] 
 has\_hashtag1 & $-$0.046$^{***}$ (0.008) \\ 
  has\_mention1 & 0.011 (0.007) \\ 
  has\_url1 & 0.185$^{***}$ (0.010) \\ 
  is\_quote\_status1 & $-$0.221$^{***}$ (0.008) \\ 
  is\_reply1 & $-$0.229$^{***}$ (0.007) \\ 
  is\_verified1 & 0.345$^{***}$ (0.014) \\ 
  log\_chars & 0.799$^{***}$ (0.007) \\ 
  log\_followers & $-$0.060$^{***}$ (0.002) \\ 
  log\_following & 0.022$^{***}$ (0.003) \\ 
  log\_statuses & 0.077$^{***}$ (0.002) \\ 
  ideology & $-$0.043$^{***}$ (0.001) \\ 
  Constant & $-$6.462$^{***}$ (0.052) \\ 
 \hline \\[-1.8ex] 
Observations & 1,680,164 \\ 
Log Likelihood & $-$652,442.400 \\ 
Akaike Inf. Crit. & 1,304,915.000 \\ 
Bayesian Inf. Crit. & 1,305,100.000 \\ 
\hline 
\hline \\[-1.8ex] 
\textit{Note:}  & \multicolumn{1}{r}{$^{*}$p$<$0.1; $^{**}$p$<$0.05; $^{***}$p$<$0.01} \\ 
\end{tabular} }
\end{table} 

\begin{table}[!htbp] \centering 
  \caption{Morality.and.Ethics} 
  \label{} 
\resizebox{.8\columnwidth}{!}{%

\begin{tabular}{@{\extracolsep{5pt}}lc} 
\hline \\[-1.8ex] 
 has\_hashtag1 & 0.186$^{***}$ (0.009) \\ 
  has\_mention1 & $-$0.077$^{***}$ (0.008) \\ 
  has\_url1 & $-$0.013 (0.012) \\ 
  is\_quote\_status1 & 0.257$^{***}$ (0.009) \\ 
  is\_reply1 & $-$0.058$^{***}$ (0.009) \\ 
  is\_verified1 & $-$0.214$^{***}$ (0.016) \\ 
  log\_chars & 0.905$^{***}$ (0.008) \\ 
  log\_followers & 0.013$^{***}$ (0.003) \\ 
  log\_following & 0.018$^{***}$ (0.003) \\ 
  log\_statuses & $-$0.034$^{***}$ (0.002) \\ 
  ideology & $-$0.411$^{***}$ (0.002) \\ 
  Constant & $-$6.804$^{***}$ (0.082) \\ 
 \hline \\[-1.8ex] 
Observations & 1,680,164 \\ 
Log Likelihood & $-$478,969.600 \\ 
Akaike Inf. Crit. & 957,969.100 \\ 
Bayesian Inf. Crit. & 958,154.100 \\ 
\hline 
\hline \\[-1.8ex] 
\textit{Note:}  & \multicolumn{1}{r}{$^{*}$p$<$0.1; $^{**}$p$<$0.05; $^{***}$p$<$0.01} \\ 
\end{tabular} }
\end{table} 

\begin{table}[!htbp] \centering 
  \caption{Policy.Prescription.and.Evaluation} 
  \label{} 
\resizebox{.8\columnwidth}{!}{%

\begin{tabular}{@{\extracolsep{5pt}}lc} 
\hline \\[-1.8ex] 
 has\_hashtag1 & 0.223$^{***}$ (0.006) \\ 
  has\_mention1 & 0.086$^{***}$ (0.005) \\ 
  has\_url1 & 0.126$^{***}$ (0.007) \\ 
  is\_quote\_status1 & $-$0.001 (0.006) \\ 
  is\_reply1 & $-$0.162$^{***}$ (0.006) \\ 
  is\_verified1 & 0.336$^{***}$ (0.011) \\ 
  log\_chars & 0.730$^{***}$ (0.005) \\ 
  log\_followers & $-$0.038$^{***}$ (0.002) \\ 
  log\_following & $-$0.011$^{***}$ (0.002) \\ 
  log\_statuses & 0.039$^{***}$ (0.001) \\ 
  ideology & 0.037$^{***}$ (0.001) \\ 
  Constant & $-$4.844$^{***}$ (0.037) \\ 
 \hline \\[-1.8ex] 
Observations & 1,680,164 \\ 
Log Likelihood & $-$1,001,011.000 \\ 
Akaike Inf. Crit. & 2,002,052.000 \\ 
Bayesian Inf. Crit. & 2,002,237.000 \\ 
\hline 
\hline \\[-1.8ex] 
\textit{Note:}  & \multicolumn{1}{r}{$^{*}$p$<$0.1; $^{**}$p$<$0.05; $^{***}$p$<$0.01} \\ 
\end{tabular} }
\end{table} 

\begin{table}[!htbp] \centering 
  \caption{Political.Factors.and.Implications} 
  \label{} 
\resizebox{.8\columnwidth}{!}{%

\begin{tabular}{@{\extracolsep{5pt}}lc} 
\hline \\[-1.8ex] 
 has\_hashtag1 & 0.267$^{***}$ (0.006) \\ 
  has\_mention1 & 0.101$^{***}$ (0.005) \\ 
  has\_url1 & 0.152$^{***}$ (0.006) \\ 
  is\_quote\_status1 & 0.113$^{***}$ (0.006) \\ 
  is\_reply1 & $-$0.333$^{***}$ (0.005) \\ 
  is\_verified1 & $-$0.186$^{***}$ (0.010) \\ 
  log\_chars & 1.063$^{***}$ (0.005) \\ 
  log\_followers & $-$0.010$^{***}$ (0.002) \\ 
  log\_following & $-$0.009$^{***}$ (0.002) \\ 
  log\_statuses & 0.064$^{***}$ (0.001) \\ 
  ideology & 0.066$^{***}$ (0.001) \\ 
  Constant & $-$6.140$^{***}$ (0.034) \\ 
 \hline \\[-1.8ex] 
Observations & 1,680,164 \\ 
Log Likelihood & $-$1,097,548.000 \\ 
Akaike Inf. Crit. & 2,195,126.000 \\ 
Bayesian Inf. Crit. & 2,195,311.000 \\ 
\hline 
\hline \\[-1.8ex] 
\textit{Note:}  & \multicolumn{1}{r}{$^{*}$p$<$0.1; $^{**}$p$<$0.05; $^{***}$p$<$0.01} \\ 
\end{tabular} }
\end{table} 

\begin{table}[!htbp] \centering 
  \caption{Public.Sentiment} 
  \label{} 
\resizebox{.8\columnwidth}{!}{%

\begin{tabular}{@{\extracolsep{5pt}}lc} 
\hline \\[-1.8ex] 
 has\_hashtag1 & 0.368$^{***}$ (0.011) \\ 
  has\_mention1 & $-$0.030$^{***}$ (0.010) \\ 
  has\_url1 & 0.383$^{***}$ (0.016) \\ 
  is\_quote\_status1 & $-$0.142$^{***}$ (0.012) \\ 
  is\_reply1 & $-$0.308$^{***}$ (0.011) \\ 
  is\_verified1 & $-$0.065$^{***}$ (0.021) \\ 
  log\_chars & 0.581$^{***}$ (0.010) \\ 
  log\_followers & 0.047$^{***}$ (0.004) \\ 
  log\_following & $-$0.020$^{***}$ (0.004) \\ 
  log\_statuses & $-$0.018$^{***}$ (0.003) \\ 
  ideology & $-$0.060$^{***}$ (0.002) \\ 
  Constant & $-$6.130$^{***}$ (0.088) \\ 
 \hline \\[-1.8ex] 
Observations & 1,680,164 \\ 
Log Likelihood & $-$332,620.700 \\ 
Akaike Inf. Crit. & 665,271.500 \\ 
Bayesian Inf. Crit. & 665,456.500 \\ 
\hline 
\hline \\[-1.8ex] 
\textit{Note:}  & \multicolumn{1}{r}{$^{*}$p$<$0.1; $^{**}$p$<$0.05; $^{***}$p$<$0.01} \\ 
\end{tabular} }
\end{table} 

\begin{table}[!htbp] \centering 
  \caption{Quality.of.Life} 
  \label{} 
\resizebox{.8\columnwidth}{!}{%

\begin{tabular}{@{\extracolsep{5pt}}lc} 

\hline \\[-1.8ex] 
 has\_hashtag1 & $-$0.159$^{***}$ (0.015) \\ 
  has\_mention1 & $-$0.452$^{***}$ (0.013) \\ 
  has\_url1 & 0.122$^{***}$ (0.018) \\ 
  is\_quote\_status1 & 0.184$^{***}$ (0.013) \\ 
  is\_reply1 & 0.366$^{***}$ (0.013) \\ 
  is\_verified1 & 0.046$^{**}$ (0.022) \\ 
  log\_chars & 1.071$^{***}$ (0.012) \\ 
  log\_followers & 0.069$^{***}$ (0.004) \\ 
  log\_following & $-$0.039$^{***}$ (0.004) \\ 
  log\_statuses & $-$0.100$^{***}$ (0.003) \\ 
  ideology & $-$0.103$^{***}$ (0.002) \\ 
  Constant & $-$7.934$^{***}$ (0.070) \\ 
 \hline \\[-1.8ex] 
Observations & 1,680,164 \\ 
Log Likelihood & $-$287,794.400 \\ 
Akaike Inf. Crit. & 575,618.700 \\ 
Bayesian Inf. Crit. & 575,803.700 \\ 
\hline 
\hline \\[-1.8ex] 
\textit{Note:}  & \multicolumn{1}{r}{$^{*}$p$<$0.1; $^{**}$p$<$0.05; $^{***}$p$<$0.01} \\ 
\end{tabular} }
\end{table} 

\begin{table}[!htbp] \centering 
  \caption{Security.and.Defense} 
  \label{} 
\resizebox{.8\columnwidth}{!}{%

\begin{tabular}{@{\extracolsep{5pt}}lc} 
\hline \\[-1.8ex] 
 has\_hashtag1 & 0.306$^{***}$ (0.007) \\ 
  has\_mention1 & $-$0.075$^{***}$ (0.006) \\ 
  has\_url1 & 0.254$^{***}$ (0.008) \\ 
  is\_quote\_status1 & $-$0.075$^{***}$ (0.006) \\ 
  is\_reply1 & $-$0.142$^{***}$ (0.006) \\ 
  is\_verified1 & $-$0.043$^{***}$ (0.013) \\ 
  log\_chars & 0.697$^{***}$ (0.006) \\ 
  log\_followers & $-$0.013$^{***}$ (0.002) \\ 
  log\_following & $-$0.014$^{***}$ (0.002) \\ 
  log\_statuses & 0.023$^{***}$ (0.002) \\ 
  ideology & 0.177$^{***}$ (0.001) \\ 
  Constant & $-$5.359$^{***}$ (0.080) \\ 
 \hline \\[-1.8ex] 
Observations & 1,680,164 \\ 
Log Likelihood & $-$859,551.100 \\ 
Akaike Inf. Crit. & 1,719,132.000 \\ 
Bayesian Inf. Crit. & 1,719,317.000 \\ 
\hline 
\hline \\[-1.8ex] 
\textit{Note:}  & \multicolumn{1}{r}{$^{*}$p$<$0.1; $^{**}$p$<$0.05; $^{***}$p$<$0.01} \\ 
\end{tabular} }
\end{table}

\begin{table}[!htbp] \centering 
  \caption{Threat..Fiscal} 
  \label{}
\resizebox{.8\columnwidth}{!}{%

\begin{tabular}{@{\extracolsep{5pt}}lc} 

\hline \\[-1.8ex] 
 has\_hashtag1 & $-$0.157$^{***}$ (0.012) \\ 
  has\_mention1 & $-$0.0002 (0.011) \\ 
  has\_url1 & $-$0.279$^{***}$ (0.011) \\ 
  is\_quote\_status1 & 0.586$^{***}$ (0.010) \\ 
  is\_reply1 & 0.573$^{***}$ (0.011) \\ 
  is\_verified1 & $-$1.036$^{***}$ (0.039) \\ 
  log\_chars & 0.576$^{***}$ (0.009) \\ 
  log\_followers & $-$0.037$^{***}$ (0.004) \\ 
  log\_following & 0.021$^{***}$ (0.004) \\ 
  log\_statuses & $-$0.037$^{***}$ (0.002) \\ 
  ideology & 0.505$^{***}$ (0.002) \\ 
  Constant & $-$5.913$^{***}$ (0.098) \\ 
 \hline \\[-1.8ex] 
Observations & 1,680,164 \\ 
Log Likelihood & $-$440,199.100 \\ 
Akaike Inf. Crit. & 880,428.300 \\ 
Bayesian Inf. Crit. & 880,613.300 \\ 
\hline 
\hline \\[-1.8ex] 
\textit{Note:}  & \multicolumn{1}{r}{$^{*}$p$<$0.1; $^{**}$p$<$0.05; $^{***}$p$<$0.01} \\ 
\end{tabular} }
\end{table} 

\begin{table}[!htbp] \centering 
  \caption{Threat..Jobs} 
  \label{} 
\resizebox{.8\columnwidth}{!}{%

\begin{tabular}{@{\extracolsep{5pt}}lc} 
\hline \\[-1.8ex] 
 has\_hashtag1 & $-$0.061$^{*}$ (0.035) \\ 
  has\_mention1 & $-$0.170$^{***}$ (0.031) \\ 
  has\_url1 & $-$0.220$^{***}$ (0.033) \\ 
  is\_quote\_status1 & 0.465$^{***}$ (0.031) \\ 
  is\_reply1 & 0.693$^{***}$ (0.032) \\ 
  is\_verified1 & $-$0.649$^{***}$ (0.091) \\ 
  log\_chars & 0.610$^{***}$ (0.026) \\ 
  log\_followers & $-$0.097$^{***}$ (0.010) \\ 
  log\_following & $-$0.011 (0.011) \\ 
  log\_statuses & 0.037$^{***}$ (0.007) \\ 
  ideology & 0.094$^{***}$ (0.005) \\ 
  Constant & $-$7.864$^{***}$ (0.136) \\ 
 \hline \\[-1.8ex] 
Observations & 1,680,164 \\ 
Log Likelihood & $-$80,905.320 \\ 
Akaike Inf. Crit. & 161,840.600 \\ 
Bayesian Inf. Crit. & 162,025.700 \\ 
\hline 
\hline \\[-1.8ex] 
\textit{Note:}  & \multicolumn{1}{r}{$^{*}$p$<$0.1; $^{**}$p$<$0.05; $^{***}$p$<$0.01} \\ 
\end{tabular} }
\end{table} 

\begin{table}[!htbp] \centering 
  \caption{Threat..National.Cohesion} 
  \label{} 
\resizebox{.8\columnwidth}{!}{%
\begin{tabular}{@{\extracolsep{5pt}}lc} 
\hline \\[-1.8ex] 
 has\_hashtag1 & 0.270$^{***}$ (0.011) \\ 
  has\_mention1 & $-$0.139$^{***}$ (0.011) \\ 
  has\_url1 & $-$0.451$^{***}$ (0.012) \\ 
  is\_quote\_status1 & 0.756$^{***}$ (0.011) \\ 
  is\_reply1 & 0.634$^{***}$ (0.012) \\ 
  is\_verified1 & $-$1.171$^{***}$ (0.043) \\ 
  log\_chars & 0.806$^{***}$ (0.010) \\ 
  log\_followers & $-$0.016$^{***}$ (0.004) \\ 
  log\_following & 0.017$^{***}$ (0.004) \\ 
  log\_statuses & $-$0.009$^{***}$ (0.003) \\ 
  ideology & 0.479$^{***}$ (0.002) \\ 
  Constant & $-$7.435$^{***}$ (0.066) \\ 
 \hline \\[-1.8ex] 
Observations & 1,680,164 \\ 
Log Likelihood & $-$382,405.800 \\ 
Akaike Inf. Crit. & 764,841.500 \\ 
Bayesian Inf. Crit. & 765,026.600 \\ 
\hline 
\hline \\[-1.8ex] 
\textit{Note:}  & \multicolumn{1}{r}{$^{*}$p$<$0.1; $^{**}$p$<$0.05; $^{***}$p$<$0.01} \\ 
\end{tabular} }
\end{table} 

\begin{table}[!htbp] \centering 
  \caption{Threat..Public.Order} 
  \label{} 
\resizebox{.8\columnwidth}{!}{%

\begin{tabular}{@{\extracolsep{5pt}}lc} 

\hline \\[-1.8ex] 
 has\_hashtag1 & 0.190$^{***}$ (0.007) \\ 
  has\_mention1 & $-$0.012$^{**}$ (0.006) \\ 
  has\_url1 & $-$0.005 (0.007) \\ 
  is\_quote\_status1 & 0.314$^{***}$ (0.006) \\ 
  is\_reply1 & 0.258$^{***}$ (0.006) \\ 
  is\_verified1 & $-$0.666$^{***}$ (0.016) \\ 
  log\_chars & 0.421$^{***}$ (0.005) \\ 
  log\_followers & $-$0.042$^{***}$ (0.002) \\ 
  log\_following & 0.035$^{***}$ (0.002) \\ 
  log\_statuses & $-$0.005$^{***}$ (0.001) \\ 
  ideology & 0.506$^{***}$ (0.001) \\ 
  Constant & $-$3.815$^{***}$ (0.054) \\ 
 \hline \\[-1.8ex] 
Observations & 1,680,164 \\ 
Log Likelihood & $-$902,962.200 \\ 
Akaike Inf. Crit. & 1,805,954.000 \\ 
Bayesian Inf. Crit. & 1,806,139.000 \\ 
\hline 
\hline \\[-1.8ex] 
\textit{Note:}  & \multicolumn{1}{r}{$^{*}$p$<$0.1; $^{**}$p$<$0.05; $^{***}$p$<$0.01} \\ 
\end{tabular} }
\end{table}

\begin{table}[!htbp] \centering 
  \caption{Victim..Discrimination} 
  \label{} 
\resizebox{.8\columnwidth}{!}{%

\begin{tabular}{@{\extracolsep{5pt}}lc} 
\hline \\[-1.8ex] 
 has\_hashtag1 & $-$0.199$^{***}$ (0.011) \\ 
  has\_mention1 & 0.180$^{***}$ (0.009) \\ 
  has\_url1 & $-$0.213$^{***}$ (0.013) \\ 
  is\_quote\_status1 & 0.232$^{***}$ (0.010) \\ 
  is\_reply1 & $-$0.003 (0.009) \\ 
  is\_verified1 & 0.300$^{***}$ (0.018) \\ 
  log\_chars & 1.120$^{***}$ (0.009) \\ 
  log\_followers & $-$0.151$^{***}$ (0.003) \\ 
  log\_following & 0.124$^{***}$ (0.004) \\ 
  log\_statuses & 0.111$^{***}$ (0.002) \\ 
  ideology & $-$0.485$^{***}$ (0.002) \\ 
  Constant & $-$8.942$^{***}$ (0.072) \\ 
 \hline \\[-1.8ex] 
Observations & 1,680,164 \\ 
Log Likelihood & $-$417,485.600 \\ 
Akaike Inf. Crit. & 835,001.200 \\ 
Bayesian Inf. Crit. & 835,186.200 \\ 
\hline 
\hline \\[-1.8ex] 
\textit{Note:}  & \multicolumn{1}{r}{$^{*}$p$<$0.1; $^{**}$p$<$0.05; $^{***}$p$<$0.01} \\ 
\end{tabular} }
\end{table} 

\begin{table}[!htbp] \centering 
  \caption{Victim..Global.Economy} 
  \label{} 
  \resizebox{.8\columnwidth}{!}{%
\begin{tabular}{@{\extracolsep{5pt}}lc} 
\hline \\[-1.8ex] 
 has\_hashtag1 & $-$0.387$^{***}$ (0.051) \\ 
  has\_mention1 & $-$0.037 (0.039) \\ 
  has\_url1 & 0.033 (0.051) \\ 
  is\_quote\_status1 & $-$0.078$^{*}$ (0.044) \\ 
  is\_reply1 & 0.156$^{***}$ (0.039) \\ 
  is\_verified1 & 0.445$^{***}$ (0.071) \\ 
  log\_chars & 0.876$^{***}$ (0.035) \\ 
  log\_followers & $-$0.065$^{***}$ (0.012) \\ 
  log\_following & 0.006 (0.014) \\ 
  log\_statuses & $-$0.024$^{***}$ (0.009) \\ 
  ideology & $-$0.053$^{***}$ (0.007) \\ 
  Constant & $-$9.659$^{***}$ (0.194) \\ 
 \hline \\[-1.8ex] 
Observations & 1,680,164 \\ 
Log Likelihood & $-$47,175.170 \\ 
Akaike Inf. Crit. & 94,380.340 \\ 
Bayesian Inf. Crit. & 94,565.350 \\ 
\hline 
\hline \\[-1.8ex] 
\textit{Note:}  & \multicolumn{1}{r}{$^{*}$p$<$0.1; $^{**}$p$<$0.05; $^{***}$p$<$0.01} \\ 
\end{tabular} }
\end{table} 

\begin{table}[!htbp] \centering 
  \caption{Victim..Humanitarian} 
  \label{} 
 \resizebox{.8\columnwidth}{!}{%
\begin{tabular}{@{\extracolsep{5pt}}lc} 
\hline \\[-1.8ex] 
 has\_hashtag1 & 0.184$^{***}$ (0.009) \\ 
  has\_mention1 & $-$0.168$^{***}$ (0.008) \\ 
  has\_url1 & 0.460$^{***}$ (0.013) \\ 
  is\_quote\_status1 & $-$0.177$^{***}$ (0.010) \\ 
  is\_reply1 & $-$0.506$^{***}$ (0.009) \\ 
  is\_verified1 & 0.188$^{***}$ (0.015) \\ 
  log\_chars & 0.491$^{***}$ (0.008) \\ 
  log\_followers & 0.005 (0.003) \\ 
  log\_following & 0.035$^{***}$ (0.003) \\ 
  log\_statuses & 0.021$^{***}$ (0.002) \\ 
  ideology & $-$0.603$^{***}$ (0.002) \\ 
  Constant & $-$5.537$^{***}$ (0.103) \\ 
 \hline \\[-1.8ex] 
Observations & 1,680,164 \\ 
Log Likelihood & $-$409,562.800 \\ 
Akaike Inf. Crit. & 819,155.600 \\ 
Bayesian Inf. Crit. & 819,340.600 \\ 
\hline 
\hline \\[-1.8ex] 
\textit{Note:}  & \multicolumn{1}{r}{$^{*}$p$<$0.1; $^{**}$p$<$0.05; $^{***}$p$<$0.01} \\ 
\end{tabular} }
\end{table} 

\begin{table}[!htbp] \centering 
  \caption{Victim..War} 
  \label{} 
\resizebox{.8\columnwidth}{!}{%
\begin{tabular}{@{\extracolsep{5pt}}lc} 

\hline \\[-1.8ex] 
 has\_hashtag1 & $-$0.067 (0.052) \\ 
  has\_mention1 & $-$0.468$^{***}$ (0.045) \\ 
  has\_url1 & 0.441$^{***}$ (0.078) \\ 
  is\_quote\_status1 & $-$0.116$^{**}$ (0.050) \\ 
  is\_reply1 & 0.173$^{***}$ (0.047) \\ 
  is\_verified1 & 0.379$^{***}$ (0.074) \\ 
  log\_chars & 1.010$^{***}$ (0.044) \\ 
  log\_followers & $-$0.021 (0.014) \\ 
  log\_following & 0.0003 (0.016) \\ 
  log\_statuses & $-$0.004 (0.011) \\ 
  ideology & $-$0.465$^{***}$ (0.010) \\ 
  Constant & $-$11.222$^{***}$ (0.240) \\ 
 \hline \\[-1.8ex] 
Observations & 1,680,164 \\ 
Log Likelihood & $-$31,227.890 \\ 
Akaike Inf. Crit. & 62,485.790 \\ 
Bayesian Inf. Crit. & 62,670.810 \\ 
\hline 
\hline \\[-1.8ex] 
\textit{Note:}  & \multicolumn{1}{r}{$^{*}$p$<$0.1; $^{**}$p$<$0.05; $^{***}$p$<$0.01} \\ 
\end{tabular} }
\end{table} 

\begin{table}[!htbp] \centering 
  \caption{Hero..Cultural.Diversity} 
  \label{} 
\resizebox{.8\columnwidth}{!}{%

\begin{tabular}{@{\extracolsep{5pt}}lc} 
\hline \\[-1.8ex] 
 has\_hashtag1 & 0.209$^{***}$ (0.026) \\ 
  has\_mention1 & $-$0.067$^{***}$ (0.025) \\ 
  has\_url1 & $-$0.562$^{***}$ (0.029) \\ 
  is\_quote\_status1 & 0.448$^{***}$ (0.025) \\ 
  is\_reply1 & 0.343$^{***}$ (0.026) \\ 
  is\_verified1 & 0.184$^{***}$ (0.043) \\ 
  log\_chars & 0.431$^{***}$ (0.023) \\ 
  log\_followers & 0.040$^{***}$ (0.008) \\ 
  log\_following & 0.001 (0.009) \\ 
  log\_statuses & $-$0.133$^{***}$ (0.006) \\ 
  ideology & $-$0.533$^{***}$ (0.006) \\ 
  Constant & $-$5.399$^{***}$ (0.141) \\ 
 \hline \\[-1.8ex] 
Observations & 1,680,164 \\ 
Log Likelihood & $-$89,983.880 \\ 
Akaike Inf. Crit. & 179,997.800 \\ 
Bayesian Inf. Crit. & 180,182.800 \\ 
\hline 
\hline \\[-1.8ex] 
\textit{Note:}  & \multicolumn{1}{r}{$^{*}$p$<$0.1; $^{**}$p$<$0.05; $^{***}$p$<$0.01} \\ 
\end{tabular} }
\end{table} 

\begin{table}[!htbp] \centering 
  \caption{Hero..Integration} 
  \label{} 
\resizebox{.8\columnwidth}{!}{%

\begin{tabular}{@{\extracolsep{5pt}}lc} 
\hline \\[-1.8ex] 
 has\_hashtag1 & $-$0.076$^{***}$ (0.020) \\ 
  has\_mention1 & $-$0.244$^{***}$ (0.017) \\ 
  has\_url1 & $-$0.158$^{***}$ (0.022) \\ 
  is\_quote\_status1 & 0.188$^{***}$ (0.018) \\ 
  is\_reply1 & 0.235$^{***}$ (0.018) \\ 
  is\_verified1 & 0.205$^{***}$ (0.029) \\ 
  log\_chars & 0.636$^{***}$ (0.016) \\ 
  log\_followers & 0.069$^{***}$ (0.005) \\ 
  log\_following & $-$0.036$^{***}$ (0.006) \\ 
  log\_statuses & $-$0.122$^{***}$ (0.004) \\ 
  ideology & $-$0.297$^{***}$ (0.003) \\ 
  Constant & $-$5.862$^{***}$ (0.093) \\ 
 \hline \\[-1.8ex] 
Observations & 1,680,164 \\ 
Log Likelihood & $-$166,396.500 \\ 
Akaike Inf. Crit. & 332,823.000 \\ 
Bayesian Inf. Crit. & 333,008.000 \\ 
\hline 
\hline \\[-1.8ex] 
\textit{Note:}  & \multicolumn{1}{r}{$^{*}$p$<$0.1; $^{**}$p$<$0.05; $^{***}$p$<$0.01} \\ 
\end{tabular} }
\end{table} 

\begin{table}[!htbp] \centering 
  \caption{Hero..Worker} 
  \label{} 
\resizebox{.8\columnwidth}{!}{%

\begin{tabular}{@{\extracolsep{5pt}}lc} 
\hline \\[-1.8ex] 
 has\_hashtag1 & $-$0.246$^{***}$ (0.031) \\ 
  has\_mention1 & $-$0.092$^{***}$ (0.025) \\ 
  has\_url1 & $-$0.027 (0.029) \\ 
  is\_quote\_status1 & 0.166$^{***}$ (0.028) \\ 
  is\_reply1 & 0.600$^{***}$ (0.026) \\ 
  is\_verified1 & 0.091$^{**}$ (0.046) \\ 
  log\_chars & 0.348$^{***}$ (0.022) \\ 
  log\_followers & $-$0.016$^{**}$ (0.008) \\ 
  log\_following & $-$0.004 (0.009) \\ 
  log\_statuses & $-$0.048$^{***}$ (0.006) \\ 
  ideology & $-$0.322$^{***}$ (0.005) \\ 
  Constant & $-$5.766$^{***}$ (0.136) \\ 
 \hline \\[-1.8ex] 
Observations & 1,680,164 \\ 
Log Likelihood & $-$103,008.500 \\ 
Akaike Inf. Crit. & 206,047.000 \\ 
Bayesian Inf. Crit. & 206,232.000 \\ 
\hline 
\hline \\[-1.8ex] 
\textit{Note:}  & \multicolumn{1}{r}{$^{*}$p$<$0.1; $^{**}$p$<$0.05; $^{***}$p$<$0.01} \\ 
\end{tabular} }
\end{table}

\begin{table}[!htbp] \centering 
  \caption{Episodic} 
  \label{} 
\resizebox{.8\columnwidth}{!}{%

\begin{tabular}{@{\extracolsep{5pt}}lc} 
\hline \\[-1.8ex] 
 has\_hashtag1 & 0.038$^{***}$ (0.007) \\ 
  has\_mention1 & 0.263$^{***}$ (0.006) \\ 
  has\_url1 & 0.560$^{***}$ (0.006) \\ 
  is\_quote\_status1 & $-$0.969$^{***}$ (0.006) \\ 
  is\_reply1 & $-$1.188$^{***}$ (0.006) \\ 
  is\_verified1 & 0.395$^{***}$ (0.014) \\ 
  log\_chars & $-$0.369$^{***}$ (0.005) \\ 
  log\_followers & 0.018$^{***}$ (0.002) \\ 
  log\_following & 0.024$^{***}$ (0.002) \\ 
  log\_statuses & 0.033$^{***}$ (0.001) \\ 
  ideology & $-$0.171$^{***}$ (0.001) \\ 
  Constant & 2.346$^{***}$ (0.040) \\ 
 \hline \\[-1.8ex] 
Observations & 1,680,164 \\ 
Log Likelihood & $-$988,528.600 \\ 
Akaike Inf. Crit. & 1,977,087.000 \\ 
Bayesian Inf. Crit. & 1,977,272.000 \\ 
\hline 
\hline \\[-1.8ex] 
\textit{Note:}  & \multicolumn{1}{r}{$^{*}$p$<$0.1; $^{**}$p$<$0.05; $^{***}$p$<$0.01} \\ 
\end{tabular} }
\end{table} 

\begin{table}[!htbp] \centering 
  \caption{Thematic} 
  \label{tab:reg_ideology_end} 
\resizebox{.8\columnwidth}{!}{%

\begin{tabular}{@{\extracolsep{5pt}}lc} 

\hline \\[-1.8ex] 
 has\_hashtag1 & $-$0.170$^{***}$ (0.006) \\ 
  has\_mention1 & $-$0.164$^{***}$ (0.006) \\ 
  has\_url1 & $-$0.579$^{***}$ (0.006) \\ 
  is\_quote\_status1 & 1.069$^{***}$ (0.006) \\ 
  is\_reply1 & 1.068$^{***}$ (0.006) \\ 
  is\_verified1 & $-$0.406$^{***}$ (0.011) \\ 
  log\_chars & 2.002$^{***}$ (0.006) \\ 
  log\_followers & $-$0.013$^{***}$ (0.002) \\ 
  log\_following & $-$0.036$^{***}$ (0.002) \\ 
  log\_statuses & $-$0.045$^{***}$ (0.001) \\ 
  ideology & 0.205$^{***}$ (0.001) \\ 
  Constant & $-$9.476$^{***}$ (0.042) \\ 
 \hline \\[-1.8ex] 
Observations & 1,680,164 \\ 
Log Likelihood & $-$914,986.900 \\ 
Akaike Inf. Crit. & 1,830,004.000 \\ 
Bayesian Inf. Crit. & 1,830,189.000 \\ 
\hline 
\hline \\[-1.8ex] 
\textit{Note:}  & \multicolumn{1}{r}{$^{*}$p$<$0.1; $^{**}$p$<$0.05; $^{***}$p$<$0.01} \\ 
\end{tabular} }
\end{table} 

\onecolumn
\section{Frame-setting (audience response) regressions}
\label{app:frame-setting-regressions}
Tables \ref{tab:reg_setting_full}-\ref{tab:reg_setting_narrative} show independent variable coefficients for linear regressions predicting favorite and retweet counts from frames. Results from Table \ref{tab:reg_setting_full} are discussed in the main paper. We find comparable effects of frames on audience responses when excluding frames from the regression model with F1 scores below 0.5 (Figure \ref{tab:reg_setting_remove5}), and when fitting separate regressions for each frame type (Figures \ref{tab:reg_setting_generic}-\ref{tab:reg_setting_narrative}).

\begin{table}[!htbp] 
    \centering
  \caption{Fit audience response variables on all frames} 
    \label{tab:reg_setting_full} 

 \resizebox{.75\textwidth}{!}{%
\begin{tabular}{@{\extracolsep{5pt}}lcc} 
\\[-1.8ex] & log\_favorites & log\_retweets \\ 
\hline \\[-1.8ex] 
 Capacity.and.Resources1 & $-$0.017$^{**}$ (0.004) & $-$0.010$^{***}$ (0.003) \\ 
  Crime.and.Punishment1 & $-$0.018$^{***}$ (0.002) & 0.018$^{***}$ (0.002) \\ 
  Cultural.Identity1 & 0.020 (0.003) & 0.001 (0.002) \\ 
  Economic1 & 0.010$^{***}$ (0.002) & 0.017$^{***}$ (0.002) \\ 
  Episodic1 & 0.002$^{***}$ (0.002) & 0.013$^{***}$ (0.002) \\ 
  External.Regulation.and.Reputation1 & $-$0.020 (0.004) & $-$0.001 (0.003) \\ 
  Fairness.and.Equality1 & 0.032$^{***}$ (0.003) & 0.021$^{***}$ (0.002) \\ 
  Health.and.Safety1 & $-$0.007$^{***}$ (0.002) & 0.017$^{***}$ (0.002) \\ 
  Hero..Cultural.Diversity1 & 0.043 (0.008) & $-$0.011 (0.006) \\ 
  Hero..Integration1 & 0.100$^{***}$ (0.005) & 0.019$^{***}$ (0.004) \\ 
  Hero..Worker1 & $-$0.017 (0.007) & 0.005 (0.006) \\ 
  Legality..Constitutionality..Jurisdiction1 & 0.002$^{***}$ (0.002) & 0.020$^{***}$ (0.002) \\ 
  Morality.and.Ethics1 & 0.066$^{***}$ (0.003) & 0.022$^{***}$ (0.002) \\ 
  Policy.Prescription.and.Evaluation1 & $-$0.003$^{*}$ (0.002) & 0.004$^{***}$ (0.001) \\ 
  Political.Factors.and.Implications1 & 0.008$^{***}$ (0.002) & 0.031$^{***}$ (0.001) \\ 
  Public.Sentiment1 & 0.030$^{***}$ (0.004) & 0.023$^{***}$ (0.003) \\ 
  Quality.of.Life1 & 0.018 (0.004) & $-$0.004 (0.003) \\ 
  Security.and.Defense1 & $-$0.017$^{***}$ (0.002) & 0.013$^{***}$ (0.002) \\ 
  Thematic1 & 0.007$^{***}$ (0.002) & 0.014$^{***}$ (0.002) \\ 
  Threat..Fiscal1 & $-$0.016$^{***}$ (0.003) & 0.011$^{***}$ (0.003) \\ 
  Threat..Jobs1 & 0.019$^{***}$ (0.009) & 0.027$^{***}$ (0.007) \\ 
  Threat..National.Cohesion1 & $-$0.016$^{***}$ (0.003) & 0.013$^{***}$ (0.003) \\ 
  Threat..Public.Order1 & 0.020$^{***}$ (0.002) & 0.028$^{***}$ (0.002) \\ 
  Victim..Discrimination1 & 0.048$^{***}$ (0.004) & 0.029$^{***}$ (0.003) \\ 
  Victim..Global.Economy1 & $-$0.033 (0.012) & $-$0.014 (0.010) \\ 
  Victim..Humanitarian1 & 0.008$^{***}$ (0.003) & 0.064$^{***}$ (0.003) \\ 
  Victim..War1 & 0.013 (0.014) & 0.011 (0.012) \\ 
  has\_hashtag1 & $-$0.048$^{***}$ (0.003) & $-$0.020$^{***}$ (0.002) \\ 
  has\_mention1 & $-$0.116$^{***}$ (0.002) & $-$0.096$^{***}$ (0.002) \\ 
  has\_url1 & $-$0.299$^{***}$ (0.003) & $-$0.168$^{***}$ (0.002) \\ 
  is\_quote\_status1 & $-$0.010$^{***}$ (0.003) & $-$0.071$^{***}$ (0.002) \\ 
  is\_reply1 & 0.050$^{***}$ (0.003) & $-$0.129$^{***}$ (0.002) \\ 
  is\_verified1 & 0.666$^{***}$ (0.005) & 0.590$^{***}$ (0.004) \\ 
  log\_chars & 0.326$^{***}$ (0.002) & 0.220$^{***}$ (0.002) \\ 
  log\_followers & 0.318$^{***}$ (0.001) & 0.243$^{***}$ (0.001) \\ 
  log\_following & $-$0.123$^{***}$ (0.001) & $-$0.087$^{***}$ (0.001) \\ 
  log\_statuses & $-$0.145$^{***}$ (0.001) & $-$0.098$^{***}$ (0.001) \\ 
  ideology & $-$0.046$^{***}$ (0.001) & $-$0.007$^{***}$ (0.0004) \\ 
  Constant & $-$0.747$^{***}$ (0.033) & $-$0.739$^{***}$ (0.012) \\ 
 \hline \\[-1.8ex] 
Observations & 1,262,326 & 1,262,326 \\ 
Log Likelihood & $-$1,613,009.000 & $-$1,335,368.000 \\ 
Akaike Inf. Crit. & 3,226,105.000 & 2,670,822.000 \\ 
Bayesian Inf. Crit. & 3,226,623.000 & 2,671,340.000 \\ 
\hline 
\hline \\[-1.8ex] 
\textit{Note:}  & \multicolumn{2}{r}{$^{*}$p$<$0.05; $^{**}$p$<$0.01; $^{***}$p$<$0.005} \\ 
\end{tabular} }
\end{table}

\begin{table*}[!htbp] \centering 
  \caption{Fit audience response variables on frames with test F1 > 0.5} 
    \label{tab:reg_setting_remove5} 

 \resizebox{.75\textwidth}{!}{%

\begin{tabular}{@{\extracolsep{5pt}}lcc} 
\\[-1.8ex] & log\_favorites & log\_retweets \\ 
\hline \\[-1.8ex] 
 Capacity.and.Resources1 & $-$0.019$^{***}$ (0.004) & $-$0.010$^{***}$ (0.003) \\ 
  Crime.and.Punishment1 & $-$0.019$^{***}$ (0.002) & 0.017$^{***}$ (0.002) \\ 
  Cultural.Identity1 & 0.024 (0.002) & 0.004$^{*}$ (0.002) \\ 
  Economic1 & 0.012$^{***}$ (0.002) & 0.017$^{***}$ (0.002) \\ 
  Episodic1 & 0.003$^{***}$ (0.002) & 0.013$^{***}$ (0.002) \\ 
  External.Regulation.and.Reputation1 & $-$0.024 (0.004) & $-$0.001 (0.003) \\ 
  Fairness.and.Equality1 & 0.028$^{***}$ (0.003) & 0.021$^{***}$ (0.002) \\ 
  Health.and.Safety1 & $-$0.008$^{***}$ (0.002) & 0.016$^{***}$ (0.002) \\ 
  Hero..Cultural.Diversity1 & 0.055 (0.008) & $-$0.011 (0.006) \\ 
  Hero..Worker1 & $-$0.010 (0.007) & 0.005 (0.006) \\ 
  Legality..Constitutionality..Jurisdiction1 & 0.002$^{***}$ (0.002) & 0.019$^{***}$ (0.002) \\ 
  Morality.and.Ethics1 & 0.066$^{***}$ (0.003) & 0.021$^{***}$ (0.002) \\ 
  Policy.Prescription.and.Evaluation1 & $-$0.003$^{*}$ (0.002) & 0.004$^{***}$ (0.001) \\ 
  Political.Factors.and.Implications1 & 0.007$^{***}$ (0.002) & 0.031$^{***}$ (0.001) \\ 
  Quality.of.Life1 & 0.024 (0.004) & $-$0.003 (0.003) \\ 
  Security.and.Defense1 & $-$0.018$^{***}$ (0.002) & 0.013$^{***}$ (0.002) \\ 
  Thematic1 & 0.008$^{***}$ (0.002) & 0.015$^{***}$ (0.002) \\ 
  Threat..Fiscal1 & $-$0.017$^{***}$ (0.003) & 0.010$^{***}$ (0.003) \\ 
  Threat..Jobs1 & 0.015$^{***}$ (0.009) & 0.026$^{***}$ (0.007) \\ 
  Threat..Public.Order1 & 0.020$^{***}$ (0.002) & 0.028$^{***}$ (0.002) \\ 
  Victim..Discrimination1 & 0.048$^{***}$ (0.003) & 0.028$^{***}$ (0.003) \\ 
  Victim..Humanitarian1 & 0.007$^{***}$ (0.003) & 0.064$^{***}$ (0.003) \\ 
  has\_hashtag1 & $-$0.047$^{***}$ (0.003) & $-$0.019$^{***}$ (0.002) \\ 
  has\_mention1 & $-$0.117$^{***}$ (0.002) & $-$0.096$^{***}$ (0.002) \\ 
  has\_url1 & $-$0.298$^{***}$ (0.003) & $-$0.168$^{***}$ (0.002) \\ 
  is\_quote\_status1 & $-$0.011$^{***}$ (0.003) & $-$0.071$^{***}$ (0.002) \\ 
  is\_reply1 & 0.050$^{***}$ (0.003) & $-$0.129$^{***}$ (0.002) \\ 
  is\_verified1 & 0.667$^{***}$ (0.005) & 0.589$^{***}$ (0.004) \\ 
  log\_chars & 0.327$^{***}$ (0.002) & 0.221$^{***}$ (0.002) \\ 
  log\_followers & 0.318$^{***}$ (0.001) & 0.243$^{***}$ (0.001) \\ 
  log\_following & $-$0.123$^{***}$ (0.001) & $-$0.087$^{***}$ (0.001) \\ 
  log\_statuses & $-$0.145$^{***}$ (0.001) & $-$0.099$^{***}$ (0.001) \\ 
  ideology & $-$0.047$^{***}$ (0.001) & $-$0.007$^{***}$ (0.0004) \\ 
  Constant & $-$0.749$^{***}$ (0.033) & $-$0.743$^{***}$ (0.012) \\ 
 \hline \\[-1.8ex] 
Observations & 1,262,326 & 1,262,326 \\ 
Log Likelihood & $-$1,613,210.000 & $-$1,335,401.000 \\ 
Akaike Inf. Crit. & 3,226,496.000 & 2,670,878.000 \\ 
Bayesian Inf. Crit. & 3,226,953.000 & 2,671,336.000 \\ 
\hline 
\hline \\[-1.8ex] 
\textit{Note:}  & \multicolumn{2}{r}{$^{*}$p$<$0.05; $^{**}$p$<$0.01; $^{***}$p$<$0.005} \\ 
\end{tabular} }
\end{table*}

\begin{table*}[!htbp] \centering 
  \caption{Fit audience response variables on only issue-generic policy frames} 
    \label{tab:reg_setting_generic} 

\resizebox{.75\textwidth}{!}{%

\begin{tabular}{@{\extracolsep{5pt}}lcc} 
\\[-1.8ex] & log\_favorites & log\_retweets \\ 
\hline \\[-1.8ex] 
 Capacity.and.Resources1 & $-$0.028$^{***}$ (0.003) & $-$0.011$^{***}$ (0.003) \\ 
  Crime.and.Punishment1 & $-$0.010$^{***}$ (0.002) & 0.030$^{***}$ (0.001) \\ 
  Cultural.Identity1 & 0.029 (0.002) & 0.001 (0.002) \\ 
  Economic1 & 0.008$^{***}$ (0.002) & 0.017$^{***}$ (0.002) \\ 
  External.Regulation.and.Reputation1 & $-$0.024 (0.004) & $-$0.003 (0.003) \\ 
  Fairness.and.Equality1 & 0.048$^{***}$ (0.002) & 0.032$^{***}$ (0.002) \\ 
  Health.and.Safety1 & $-$0.006$^{***}$ (0.002) & 0.026$^{***}$ (0.002) \\ 
  Legality..Constitutionality..Jurisdiction1 & 0.003$^{***}$ (0.002) & 0.020$^{***}$ (0.002) \\ 
  Morality.and.Ethics1 & 0.069$^{***}$ (0.003) & 0.049$^{***}$ (0.002) \\ 
  Policy.Prescription.and.Evaluation1 & $-$0.003 (0.002) & 0.003$^{*}$ (0.001) \\ 
  Political.Factors.and.Implications1 & 0.006$^{***}$ (0.002) & 0.030$^{***}$ (0.001) \\ 
  Public.Sentiment1 & 0.029$^{***}$ (0.004) & 0.023$^{***}$ (0.003) \\ 
  Quality.of.Life1 & 0.023 (0.004) & 0.001 (0.003) \\ 
  Security.and.Defense1 & $-$0.013$^{***}$ (0.002) & 0.019$^{***}$ (0.002) \\ 
  has\_hashtag1 & $-$0.048$^{***}$ (0.003) & $-$0.019$^{***}$ (0.002) \\ 
  has\_mention1 & $-$0.115$^{***}$ (0.002) & $-$0.096$^{***}$ (0.002) \\ 
  has\_url1 & $-$0.300$^{***}$ (0.003) & $-$0.169$^{***}$ (0.002) \\ 
  is\_quote\_status1 & $-$0.009$^{***}$ (0.003) & $-$0.070$^{***}$ (0.002) \\ 
  is\_reply1 & 0.050$^{***}$ (0.003) & $-$0.129$^{***}$ (0.002) \\ 
  is\_verified1 & 0.667$^{***}$ (0.005) & 0.589$^{***}$ (0.004) \\ 
  log\_chars & 0.332$^{***}$ (0.002) & 0.228$^{***}$ (0.002) \\ 
  log\_followers & 0.317$^{***}$ (0.001) & 0.243$^{***}$ (0.001) \\ 
  log\_following & $-$0.122$^{***}$ (0.001) & $-$0.087$^{***}$ (0.001) \\ 
  log\_statuses & $-$0.145$^{***}$ (0.001) & $-$0.098$^{***}$ (0.001) \\ 
  ideology & $-$0.047$^{***}$ (0.0005) & $-$0.007$^{***}$ (0.0004) \\ 
  Constant & $-$0.767$^{***}$ (0.032) & $-$0.757$^{***}$ (0.011) \\ 
 \hline \\[-1.8ex] 
Observations & 1,262,326 & 1,262,326 \\ 
Log Likelihood & $-$1,613,329.000 & $-$1,335,820.000 \\ 
Akaike Inf. Crit. & 3,226,718.000 & 2,671,700.000 \\ 
Bayesian Inf. Crit. & 3,227,079.000 & 2,672,061.000 \\ 
\hline 
\hline \\[-1.8ex] 
\textit{Note:}  & \multicolumn{2}{r}{$^{*}$p$<$0.05; $^{**}$p$<$0.01; $^{***}$p$<$0.005} \\ 
\end{tabular} }
\end{table*}

\begin{table*}[!htbp] \centering 
  \caption{Fit audience response variables on only issue-specific frames} 
  \label{tab:reg_setting_specific} 

 \resizebox{.6\textwidth}{!}{%

\begin{tabular}{@{\extracolsep{5pt}}lcc} 
\\[-1.8ex] & log\_favorites & log\_retweets \\ 
\hline \\[-1.8ex] 
 Hero..Cultural.Diversity1 & 0.060$^{***}$ (0.008) & $-$0.025$^{***}$ (0.006) \\ 
  Hero..Integration1 & 0.111 (0.005) & 0.007 (0.004) \\ 
  Hero..Worker1 & $-$0.013 (0.007) & 0.003 (0.006) \\ 
  Threat..Fiscal1 & $-$0.013$^{***}$ (0.003) & 0.014$^{***}$ (0.002) \\ 
  Threat..Jobs1 & 0.026$^{***}$ (0.009) & 0.027$^{***}$ (0.007) \\ 
  Threat..National.Cohesion1 & $-$0.003$^{***}$ (0.003) & 0.012$^{***}$ (0.003) \\ 
  Threat..Public.Order1 & 0.00003$^{***}$ (0.002) & 0.038$^{***}$ (0.002) \\ 
  Victim..Discrimination1 & 0.075$^{***}$ (0.003) & 0.040$^{***}$ (0.002) \\ 
  Victim..Global.Economy1 & $-$0.033 (0.012) & $-$0.015 (0.009) \\ 
  Victim..Humanitarian1 & 0.033$^{***}$ (0.003) & 0.078$^{***}$ (0.002) \\ 
  Victim..War1 & 0.009 (0.014) & 0.011 (0.012) \\ 
  has\_hashtag1 & $-$0.046$^{***}$ (0.003) & $-$0.018$^{***}$ (0.002) \\ 
  has\_mention1 & $-$0.117$^{***}$ (0.002) & $-$0.097$^{***}$ (0.002) \\ 
  has\_url1 & $-$0.302$^{***}$ (0.003) & $-$0.166$^{***}$ (0.002) \\ 
  is\_quote\_status1 & $-$0.005$^{***}$ (0.003) & $-$0.069$^{***}$ (0.002) \\ 
  is\_reply1 & 0.054$^{***}$ (0.003) & $-$0.130$^{***}$ (0.002) \\ 
  is\_verified1 & 0.661$^{***}$ (0.005) & 0.588$^{***}$ (0.004) \\ 
  log\_chars & 0.335$^{***}$ (0.002) & 0.243$^{***}$ (0.002) \\ 
  log\_followers & 0.318$^{***}$ (0.001) & 0.243$^{***}$ (0.001) \\ 
  log\_following & $-$0.123$^{***}$ (0.001) & $-$0.087$^{***}$ (0.001) \\ 
  log\_statuses & $-$0.146$^{***}$ (0.001) & $-$0.098$^{***}$ (0.001) \\ 
  ideology & $-$0.047$^{***}$ (0.001) & $-$0.007$^{***}$ (0.0004) \\ 
  Constant & $-$0.775$^{***}$ (0.033) & $-$0.808$^{***}$ (0.012) \\ 
 \hline \\[-1.8ex] 
Observations & 1,262,326 & 1,262,326 \\ 
Log Likelihood & $-$1,613,485.000 & $-$1,335,947.000 \\ 
Akaike Inf. Crit. & 3,227,023.000 & 2,671,949.000 \\ 
Bayesian Inf. Crit. & 3,227,349.000 & 2,672,274.000 \\ 
\hline 
\hline \\[-1.8ex] 
\textit{Note:}  & \multicolumn{2}{r}{$^{*}$p$<$0.05; $^{**}$p$<$0.01; $^{***}$p$<$0.005} \\ 
\end{tabular} }
\end{table*}

\begin{table*}[!htbp] \centering 
  \caption{Fit audience response variables on only narrative frames} 
    \label{tab:reg_setting_narrative} 

 \resizebox{.6\textwidth}{!}{%

\begin{tabular}{@{\extracolsep{5pt}}lcc} 
\\[-1.8ex] & log\_favorites & log\_retweets \\ 
\hline \\[-1.8ex] 
 Episodic1 & 0.006$^{***}$ (0.002) & 0.014$^{***}$ (0.002) \\ 
  Thematic1 & 0.014$^{***}$ (0.002) & 0.026$^{***}$ (0.002) \\ 
  has\_hashtag1 & $-$0.046$^{***}$ (0.003) & $-$0.016$^{***}$ (0.002) \\ 
  has\_mention1 & $-$0.117$^{***}$ (0.002) & $-$0.098$^{***}$ (0.002) \\ 
  has\_url1 & $-$0.303$^{***}$ (0.003) & $-$0.165$^{***}$ (0.002) \\ 
  is\_quote\_status1 & $-$0.006$^{***}$ (0.003) & $-$0.070$^{***}$ (0.002) \\ 
  is\_reply1 & 0.051$^{***}$ (0.003) & $-$0.134$^{***}$ (0.002) \\ 
  is\_verified1 & 0.665$^{***}$ (0.005) & 0.586$^{***}$ (0.004) \\ 
  log\_chars & 0.338$^{***}$ (0.002) & 0.243$^{***}$ (0.002) \\ 
  log\_followers & 0.317$^{***}$ (0.001) & 0.242$^{***}$ (0.001) \\ 
  log\_following & $-$0.122$^{***}$ (0.001) & $-$0.087$^{***}$ (0.001) \\ 
  log\_statuses & $-$0.145$^{***}$ (0.001) & $-$0.097$^{***}$ (0.001) \\ 
  ideology & $-$0.052$^{***}$ (0.0005) & $-$0.008$^{***}$ (0.0004) \\ 
  Constant & $-$0.792$^{***}$ (0.032) & $-$0.818$^{***}$ (0.012) \\ 
 \hline \\[-1.8ex] 
Observations & 1,262,326 & 1,262,326 \\ 
Log Likelihood & $-$1,614,099.000 & $-$1,336,791.000 \\ 
Akaike Inf. Crit. & 3,228,233.000 & 2,673,619.000 \\ 
Bayesian Inf. Crit. & 3,228,450.000 & 2,673,836.000 \\ 
\hline 
\hline \\[-1.8ex] 
\textit{Note:}  & \multicolumn{2}{r}{$^{*}$p$<$0.05; $^{**}$p$<$0.01; $^{***}$p$<$0.005} \\ 
\end{tabular} }
\end{table*}

\end{document}